\documentclass[]{x2lab}
\usepackage{fix-cm}
\usepackage{csquotes}
\usepackage{afterpage}
\usepackage{amsmath,amssymb,amsfonts}
\usepackage{mathtools}
\usepackage{amsthm}
\usepackage{mathrsfs}
\usepackage{xspace}
\usepackage{colortbl}
\usepackage{enumitem}
\usepackage{wrapfig}
\usepackage{makecell}
\usepackage{pifont}
\usepackage{array}
\usepackage{tabularx}
\usepackage{longtable}
\usepackage{xltabular}
\usepackage{float}
\usepackage{placeins}
\usepackage[title]{appendix}

\graphicspath{{figures/}{figs/}{Ill_Figures/}{EXP_Figures_0712/}}


\definecolor{fbApp}{HTML}{c8e7fa}
\definecolor{fbPurple3}{HTML}{f0ebf5}
\definecolor{citecolor}{HTML}{0071BC}
\definecolor{linkcolor}{HTML}{ED1C24}
\definecolor{emcolor}{HTML}{000000}
\newcommand{\myemcolor}{emcolor}

\addto\captionsenglish{}
\titleformat*{\paragraph}{\rmfamily\bfseries}

\newcommand{\method}{\textbf{HOST}\xspace}
\newcommand{\methodfull}{\textbf{HOST} (\textbf{H}uman-to-robot \textbf{O}ne-\textbf{S}hot Skill Acquisi\textbf{T}ion)\xspace}

\newcommand{\couplemech}{coupling prediction targets to the demonstration}
\newcommand{\Couplemech}{Coupling prediction targets to the demonstration}

\newcommand{\stagemech}{resolving execution through self-grounded prediction}
\newcommand{\Stagemech}{Resolving execution through self-grounded prediction}

\newcommand{\couplename}{target coupling}
\newcommand{\stagename}{self-grounded prediction}

\newcommand{\manifold}{shared task progress manifold}

\newcommand{\videoexpert}{video expert}
\newcommand{\Videoexpert}{Video expert}
\newcommand{\actionexpert}{action expert}
\newcommand{\Actionexpert}{Action expert}
\newcommand{\textenc}{UMT5-XXL}

\title{HOST: Robots Acquire Manipulation Skills in Seconds from a Single Human Video}

\author[1,\textcolor{\myemcolor}{2,\ddagger}]{Guangyan Chen}
\author[1]{Meiling Wang}
\author[1,\textcolor{\myemcolor}{2,\ddagger}]{Te Cui}
\author[1,\textcolor{\myemcolor}{2,\ddagger}]{Zichen Zhou}
\author[1]{Qi Shao}
\author[2,\dagger]{\textcolor{\myemcolor}{Xiaofan Li}}
\author[2,3]{Hang Su}
\author[2]{\textcolor{\myemcolor}{Ruyi Gan}}
\author[2,\textcolor{\myemcolor}{*}]{Hao Wang}
\author[1,*]{Mengyin Fu}
\author[1,*]{Yi Yang}
\author[1,*]{Yufeng Yue}

\affiliation[1]{Beijing Institute of Technology}
\affiliation[2]{X SQUARE ROBOT}
\affiliation[3]{Tsinghua University}
\contribution[\ddagger]{\textcolor{\myemcolor}{Work done during internships at X SQUARE ROBOT.}}
\contribution[\dagger]{Project lead.}
\contribution[*]{Corresponding author(s).}

\abstract{
The ability to acquire skills rapidly and effortlessly while retaining those already mastered is essential for robots. However, current methods still rely on a cumbersome training-time loop that is costly and slow, while eroding skills already mastered. In this paper, we introduce \methodfull, a framework that enables a robot to acquire skills in seconds from a single human video while retaining previously mastered skills.
\method resolves skill acquisition through a cascade of \stagename{}. It first estimates the robot's progress within the demonstrated task, then translates the upcoming progression into the robot's own future observations, and finally derives actions from these predicted observations. This cascade is trained on targets coupled to the video demonstration, obtained by mapping the robot trajectory and the video demonstration onto a shared task progress manifold, then redefining each target to align with the future progression of the video.
\method thereby enables the robot to actively follow the demonstrated procedure and adapt it to the robot's embodiment. \method acquires novel skills at inference time from a single human video in an average of 29 seconds and achieves a 62\% average success rate. It exceeds the zero-shot baseline by 45\% while retaining previously mastered skills. \method even exceeds the baseline fine-tuned on 50 robot demonstrations per task while requiring 50 times fewer demonstrations and acquiring each skill 507 times faster.
\textcolor{\myemcolor}{Additional information is available on the \href{https://host-site.host-robotics.workers.dev/}{project website} and \href{https://github.com/CGuangyan-BIT/HOST}{GitHub repository}.}
}

\begin{document}
\maketitle

\begin{figure}[!tp]
\centering
\includegraphics[width=\textwidth]{Ill_Figures/Fig1.pdf}
\vspace{-10pt}
\caption{\textbf{Robots acquire manipulation skills in seconds from a single human video.}
(\textbf{A})~Current methods teach a robot new skills through a cumbersome training-time loop that is costly and slow, and erodes previously mastered skills. \method instead acquires skills at inference time, each from a single human video, averaging 29 seconds, while retaining its previously mastered skills.
(\textbf{B})~Skill acquisition from a single human video confronts a structural mismatch between video demonstration and execution. Temporal asynchrony between the two decouples the prediction target from the video demonstration, while state discrepancies hinder the translation of the observed behavior into action.
(\textbf{C})~\method overcomes this mismatch through two mechanisms, \couplemech{} and \stagemech{}, enabling the robot to actively follow the demonstrated procedure and adapt it to the robot's own embodiment.}
\label{fig:overview}
\end{figure}

\section{Introduction}

Developing robots that operate seamlessly in human environments, with varied objects, and utilizing various skills to complete a broad range of tasks has been a long-standing goal in robotics. In such settings, manipulation tasks vary widely across objects, tools, procedures, and user preferences, making it infeasible to anticipate every situation a robot may encounter. 
A robot may be asked to carry out an unfamiliar household routine, adapt to a new object arrangement, or handle a tool it has never seen. The ability to acquire new skills rapidly and effortlessly while retaining those already mastered is therefore essential for robots.

Despite substantial progress in robot learning, teaching a robot each new skill still relies on a cumbersome training-time loop, a process that is costly, slow, and self-defeating. Reinforcement learning usually requires task-specific rewards and extensive trial-and-error interaction. On physical robots, such rewards are difficult to design and often unknown, while interaction remains slow, sample-inefficient and hard to scale across tasks~\cite{Hwangbo2019,Lee2020Quadruped,Chen2023VisualDexterity,OpenAI2019Rubik,Kalashnikov2018QTOpt,Kalashnikov2021MTOpt,Theodorou2010PI,Chebotar2016PIGPS,Ma2025Badminton,Luo2025HILSERL,Barreiros2025ExampleGuidedRL}. Moreover, optimizing the reward for each skill tends to degrade skills the robot has already mastered. Imitation learning offers a more intuitive and scalable alternative, enabling robots to acquire skills directly from robot demonstrations~\cite{Schaal1999,RT2,pi0,pi05,Cui2021NextGenerationManipulation}. Trained on large-scale demonstration corpora that span hundreds of tasks and diverse embodiments~\cite{DiffusionPolicy, ACT, RT1, RT2, OpenXEmbodiment, pi0, pi05, Octo}, imitation learning has markedly advanced manipulation generalization. However, the acquisition of a novel skill still relies on a training-time loop of collection and fine-tuning~\cite{Intention2Execution,RICL} that exacts a considerable price and risks eroding skills already mastered.
This loop begins with a skilled operator collecting a dedicated batch of task-specific demonstrations by teleoperating the physical robot, a slow, repeated process that consumes expert labor and scarce robot time. It then demands offline fine-tuning that incorporates these demonstrations into the policy, consumes additional compute, and extends over hours before a new skill becomes available. Even when this price is paid, the gain does not endure. Fine-tuning overwrites parameters shared across skills, causing each newly acquired behavior to erode previously mastered skills~\cite{FLaRe,CatastrophicForgettingRobot}.
This training-time loop therefore constitutes the central bottleneck for robots to acquire new skills rapidly and effortlessly while retaining those already mastered, severely hindering the deployment of robots that operate seamlessly in human environments, as depicted in Fig.~\ref{fig:overview}A.

We draw inspiration from the observational learning capabilities of humans to break through this bottleneck. When faced with an unfamiliar task, humans do not require prolonged physical rehearsal or exhaustive trial and error. Instead, they rapidly acquire new skills by observing others and adapting the observed actions to their own bodies. Motivated by this cognitive efficiency, we reformulate novel skill acquisition as an inference-time process rather than a training-time loop, in which the behaviour is specified by a single human video and translated into robot actions at inference time without task-specific parameter updates.
This reformulation supplies each new skill through a single video in place of the demonstration corpora that fine-tuning requires, yielding a pronounced improvement in data efficiency. The skill acquisition process proceeds without offline fine-tuning, enabling a new skill to become available in the time it takes to record it rather than the hours that teleoperation and offline fine-tuning would otherwise demand. Furthermore, the policy parameters remain frozen throughout the skill acquisition process, keeping previously mastered skills intact.

As early as 1994, Kuniyoshi et al.~\cite{Kuniyoshi1994} envisioned robots that acquire manipulation skills by visually observing a single human video, a setting later studied as one-shot visual imitation (OSVI). Although this vision has motivated a long line of work~\cite{FinnOSVI,DAML,MOSAIC,Vid2Robot,OSVI-WM}, achieving it remains an open challenge.
Current prevalent paradigms~\cite{FinnOSVI, DAML, MOSAIC, Vid2Robot, OSVI-WM} straightforwardly extend the imitation learning framework by conditioning on the video demonstration, while still constructing a direct mapping to robot actions and anchoring each prediction target to a fixed temporal offset along the execution trajectory.
These methods inevitably reduce video demonstrations to passive context and struggle to translate the observed procedure into precise fine-grained actions, resulting in performance substantially inferior to that attained with task-specific fine-tuning.

These persistent limitations point to a fundamental problem rooted in the structural mismatch between video demonstration and execution, illustrated in Fig.~\ref{fig:overview}B.
(1) \textbf{Temporal misalignment relegates the video demonstration to a passive context}. Video demonstrations and robot trajectories are temporally asynchronous, unfolding at disparate speeds. This decouples the prediction target from the video demonstration context, relegating the video demonstration to a passive context and hindering the model from acquiring novel skills from the video demonstration at inference time. (2) \textbf{State discrepancies between the video demonstration and robot execution hinder translation into actions}. The full video demonstration is provided as a monolithic input, flooding the model with states irrelevant to its current stage. Even at corresponding segments, substantial differences in embodiment, viewpoint, and appearance between the video demonstration and the robot trajectory introduce severe state discrepancies. These hinder direct regression from high-dimensional video demonstrations to low-dimensional robot actions. (3) \textbf{Scarce paired supervision impedes the development of reliable skill acquisition}.

Learning to acquire skills from human video for robot execution requires paired data that capture the same behavior in both human and robot demonstrations. Human videos and robot trajectories are each abundant on their own, but data pairing the two for the same behavior remains rare. This scarcity impedes the development of reliable skill acquisition from a single human video.

\begin{figure}[!tp]
  \centering
  \includegraphics[width=\textwidth,height=0.75\textheight,keepaspectratio]{Ill_Figures/Fig2.pdf}
  \caption{\textbf{Method overview of \method.}
  (\textbf{A})~{\Couplemech.}
  \method maps the robot trajectory and the video demonstration onto a shared task progress manifold, then redefines each target to align with the video's future progression, turning the video into the active driver of the robot's prediction.
  (\textbf{B})~{\Stagemech.}
  \method works through a causal cascade of self-grounded stages. It first localizes the robot's progress within the video demonstration, then translates its upcoming progression into the robot's own future observations, and finally derives actions from these predicted observations.
  (\textbf{C})~
  Given a single human video of a previously unseen task, \method runs the same self-grounded cascade to carry out the task, acquiring the skill while retaining those already mastered.}
  \label{fig:method_overview}
\end{figure}

In this paper, we introduce \methodfull, a framework for acquiring manipulation skills from a single human video demonstration, illustrated in Fig.~\ref{fig:method_overview}.
The central idea is to treat skill acquisition not as a passive, direct mapping from a human demonstration to robot action but as a process driven by the video demonstration and resolved through \stagename{}. This formulation builds on the following two principles.

First, the video demonstration should actively determine the robot's prediction target. Rather than fixing this target at a fixed temporal offset along the execution trajectory, independent of the video demonstration, we couple the target to the upcoming progression of the video demonstration, as shown in Fig.~\ref{fig:method_overview}A. This coupling is established by mapping the robot trajectory and the video demonstration onto a \manifold{}. Each target is then redefined to align with the future progression of the video, turning the video from a passive conditioning signal into the active driver of the robot's prediction.

Second, skill acquisition should be resolved through a cascade of \stagename{}.
Humans do not acquire skills from a video demonstration in a single step but instead work through a cascade of self-grounded stages. They first retrieve the task-relevant segment of the video demonstration~\cite{schacter2007prospective,meltzoff1988infant,gershman2017reinforcement}, then internally simulate the intended movement~\cite{jeannerod2001neural,wolpert2003unifying}, and finally translate that simulation into action~\cite{rizzolatti2001neurophysiological,buccino2004neural}. As illustrated in Fig.~\ref{fig:method_overview}B, \method formulates this cascade across three stages. It first localizes where the robot currently stands within the demonstrated procedure, retrieving the task-relevant segment. It then translates the upcoming progression of the video demonstration into the robot's own future observations, and finally derives actions from these predicted observations. This causal cascade bridges the gap between human video demonstration and robot action, narrowing the cross-domain discrepancy before action generation.

\method is trained in two stages to reduce its reliance on human--robot paired data. It is first trained on same-embodiment paired data, obtained by readily pairing different robot episodes of the same task from existing datasets. In this stage, the model develops the capacity to follow a demonstrated procedure by predicting future robot observations and actions conditioned on another execution of the same task. It is then adapted using a comparatively small set of human--robot paired demonstrations, extending the learned capabilities to human video demonstrations and enabling reliable skill acquisition from a single human video.

Building on the above analysis, we instantiate \method,  as illustrated in Fig.~\ref{fig:method_overview}. \method is trained on same-task pairs of video demonstrations and robot trajectories, and it then acquires novel skills at inference from a single human video. During training, each prediction target is coupled to the video demonstration, driving the robot to actively follow the demonstrated procedure. The model learns to produce this target through a cascade of \stagename{}, adapting the demonstrated procedure to the robot's own embodiment. It is trained in two stages, first on same-embodiment robot--robot pairs, then adapted to human-to-robot scenarios with a small set of human--robot pairs. After training, \method is able to acquire novel skills by running the self-grounded cascade conditioned on a single human video, without any parameter update. This video can then be stored and retrieved without repeated human involvement, enabling the acquired skill to persist across recurring tasks. Each acquisition takes 29 seconds on average, and \method reaches 62\% average success, exceeding the strongest one-shot visual imitation baseline by 43\% and the strongest zero-shot imitation learning baseline by 45\%. Even compared with the strongest baseline fine-tuned on 50 robot demonstrations per task, \method achieves superior performance while using 50 times fewer demonstrations, acquiring new skills 507 times faster, and retaining previously mastered skills.

\section{Results}

\subsection{Overview}\label{sec:overview}

\begin{figure}[!tp]
\centering
\includegraphics[width=\textwidth]{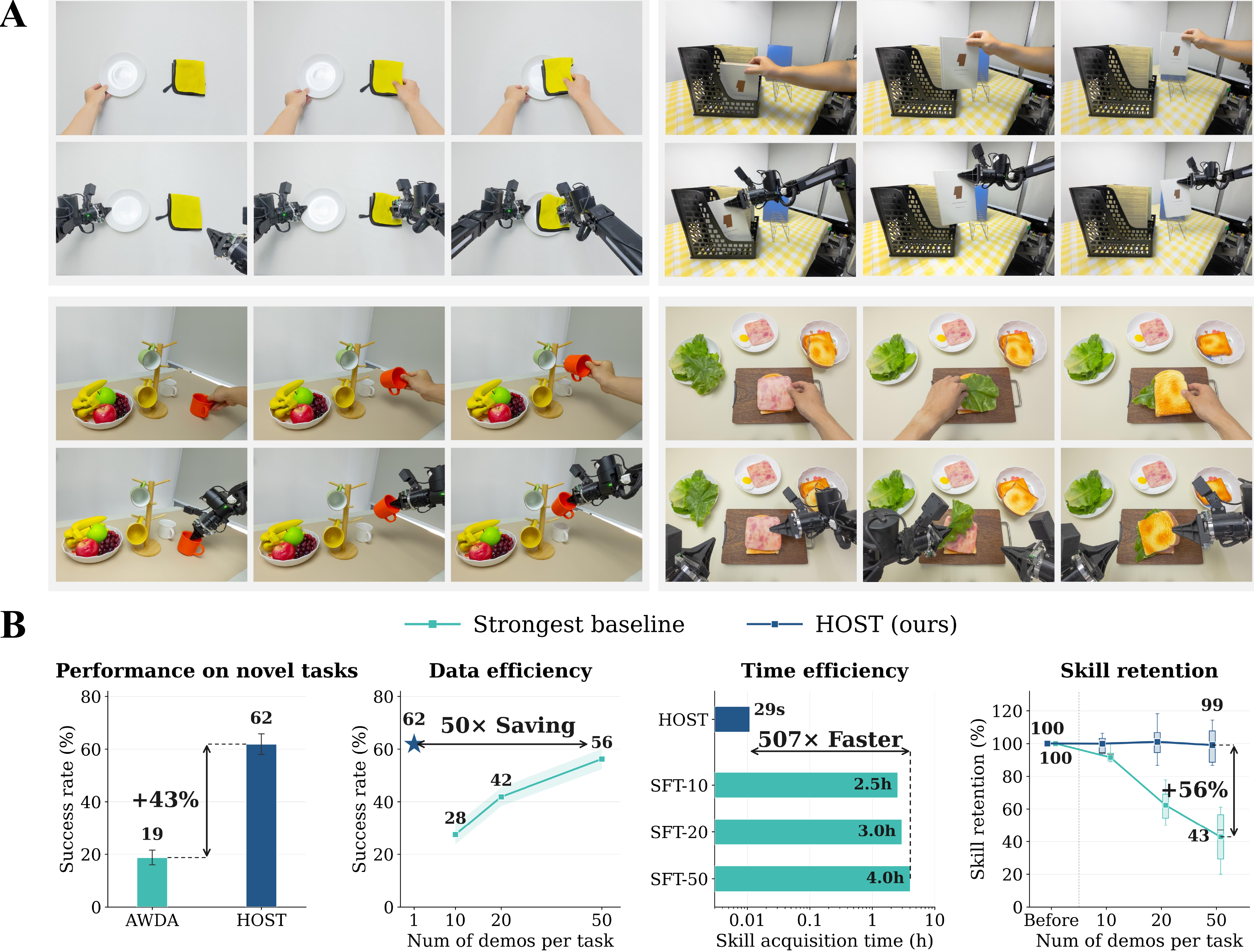}
\caption{\textbf{Qualitative and quantitative performance overview of \method.}
(\textbf{A})~\method acquires manipulation skills from a single human video, actively following the demonstrated procedure and adapting it to its own embodiment.
(\textbf{B})~\method exceeds the strongest baseline without parameter updates on novel tasks by 43\%, surpasses the strongest fine-tuned baseline while using 50 times fewer demonstrations and acquiring each new skill 507 times faster, and retains its performance on previously mastered tasks, whereas the strongest fine-tuned baseline falls to \textcolor{\myemcolor}{43\%}.}
\label{fig:results_overview}
\end{figure}

\method acquires manipulation skills from a single human video, illustrated in Fig.~\ref{fig:results_overview}A. Given the human video shown in the top row, across four representative tasks spanning varied objects, tools and manipulation primitives, \method actively follows the demonstrated procedure and adapts it to its own embodiment, carrying out each task in the bottom row from that single recording without any parameter update.

We benchmark \method against the strongest baselines in terms of performance on novel tasks, data efficiency, time efficiency, and retention of previously mastered skills, as reported in Fig.~\ref{fig:results_overview}B. \method reaches 62\% success on novel tasks from a single human video, exceeding the strongest baseline without parameter updates by 43\%, and surpasses the strongest fine-tuned baseline while using 50 times fewer demonstrations and acquiring each new skill 507 times faster. This advantage does not come at the cost of previously mastered skills. \method retains its original performance, whereas the strongest fine-tuned baseline falls to \textcolor{\myemcolor}{43\%}. The following sections examine each of these advantages in turn.

We first describe the experimental setup and baselines (Sec.~\ref{sec:setup}), then demonstrate that \method acquires skills from a single human video across 50 novel manipulation tasks (Sec.~\ref{sec:largescale}), and next compare \method against imitation learning and OSVI baselines on novel tasks without parameter updates (Sec.~\ref{sec:r1}). A further comparison against imitation learning under supervised fine-tuning measures the data and time efficiency of \method (Sec.~\ref{sec:data_eff}), and confirms that \method retains previously mastered skills where fine-tuning erodes them (Sec.~\ref{sec:r_retention}). We also assess robustness under deployment variations (Sec.~\ref{sec:r5}), then measure the contribution of \couplename{} (Sec.~\ref{sec:r2}) and the \stagename{} (Sec.~\ref{sec:r3}), and quantify the effect of same-embodiment pretraining data volume on skill acquisition from a single human video (Sec.~\ref{sec:r4}). Finally, we verify that acquired skills persist across recurring tasks, retrieved from their stored video demonstrations without repeated human involvement (Sec.~\ref{sec:r_retrieval}).

\subsection{Experimental Setup and Baselines}\label{sec:setup}

\textbf{Experimental setup.}
All experiments are conducted on a bimanual manipulation platform consisting of two ARX R5 six-axis robotic arms, each equipped with a parallel-jaw gripper. Three RGB cameras observe the workspace, with one wrist-mounted on each arm and one providing a static third-person view. Each arm operates in a 10-dimensional action space comprising Cartesian position, a 6D rotation representation, and gripper aperture, yielding a 20-dimensional bimanual action space. Complete implementation details are provided in Supplementary Sec.~\ref{sec:supp_impl}.

We evaluate skill acquisition on novel tasks that require manipulation skills absent from the training data, and assess skill retention on previously mastered tasks drawn from the training set. Each task undergoes 20 trials with randomized initial object positions and orientations, and human evaluators judge success against task-specific completion criteria.

\textbf{Baselines.}
We compare \method against baselines from two paradigms.
(1)~OSVI methods, namely Vid2Robot~\cite{Vid2Robot} and AWDA~\cite{AWDA}, are evaluated with their conditioning mechanisms integrated into the \method backbone to ensure a fair comparison. As Vid2Robot is not open source, we report results from our reimplementation. These methods receive a single human video demonstration at test time.
(2)~Imitation learning methods, namely $\pi_{0.5}$~\cite{pi05}, Wall-OSS~\cite{WallOSS}, and \method-base~\cite{FastWAM}, receive a language instruction at test time. \method-base shares the \method backbone but omits \couplename{} and \stagename{}, conditioning directly on language.

\method and the OSVI methods follow the same two-stage training protocol described in Sec.~\ref{sec:training_inference}, in which Stage~1 pretrains on same-embodiment robot--robot pairs formed from 193{,}462 robot trajectories spanning 229 tasks and Stage~2 adapts the model with human--robot pairs formed from an additional 5{,}847 self-collected human video demonstrations, each paired with robot trajectories of the same task drawn from that corpus. Among the imitation learning methods, $\pi_{0.5}$ and Wall-OSS are initialized from weights pretrained on their respective large-scale corpora and then trained on the same 193{,}462 robot trajectories, whereas \method-base is trained directly on these robot trajectories.

\subsection{Skill Acquisition across 50 Novel Manipulation Tasks}\label{sec:largescale}

\begin{figure}[!tp]
\centering
\includegraphics[width=\textwidth]{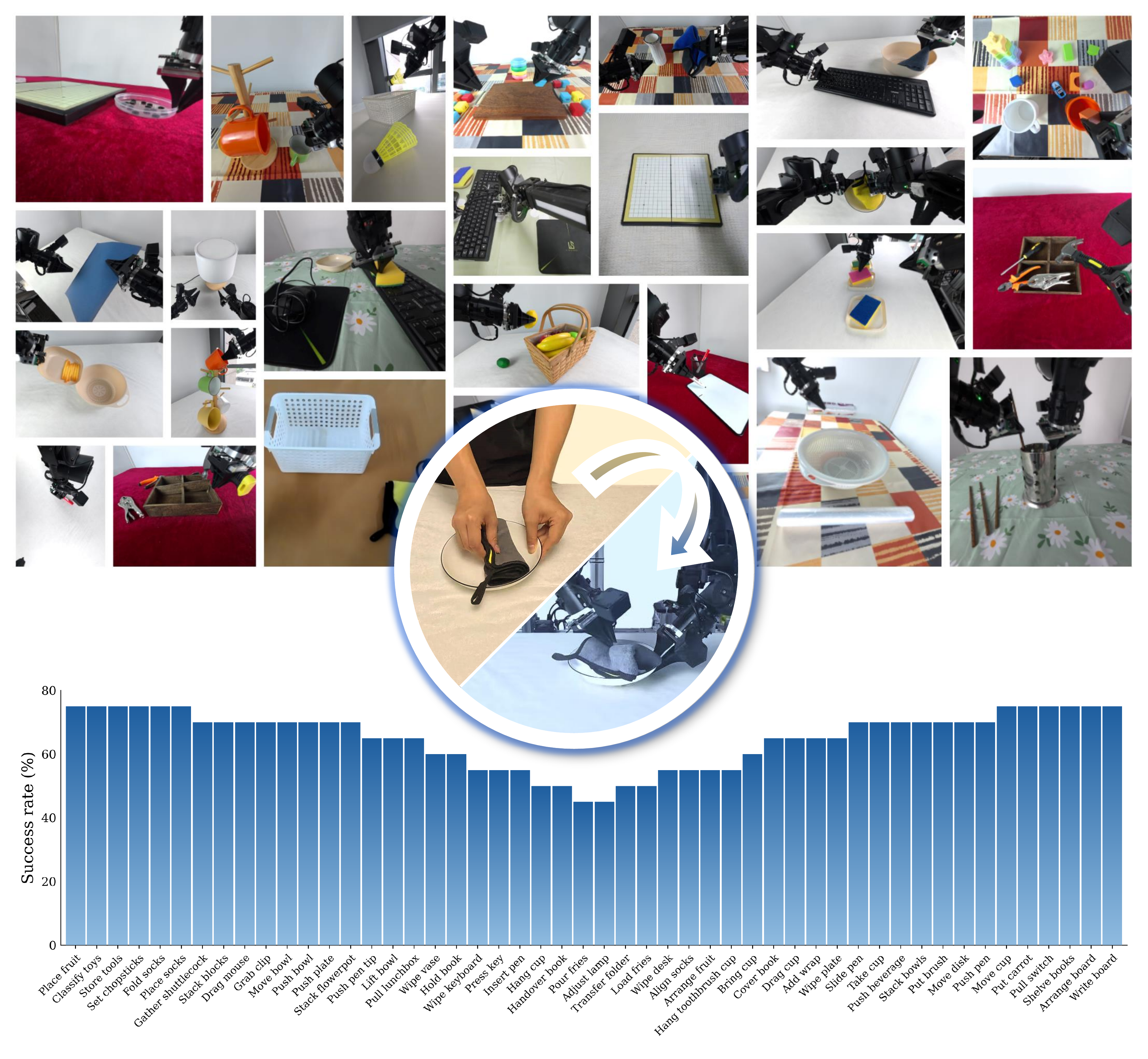}
\caption{\textbf{\method acquires manipulation skills from a single human video across 50 novel manipulation tasks.} The 50 novel manipulation tasks span varied objects, tools and manipulation primitives. \method acquires each skill from a single human video demonstration at inference time.}
\label{fig:large_scale_eval}
\end{figure}

We evaluate \method on 50 novel manipulation tasks spanning varied objects, tools, and manipulation primitives. Each task is evaluated over 20 trials, as reported in Fig.~\ref{fig:large_scale_eval}. \method acquires an executable skill on every one of these 50 tasks from only a single human video per task, without any parameter update, and sustains success across the full set. This breadth confirms that \method acquires skills from a single human video broadly across manipulation tasks. Building on this breadth, we next compare \method against imitation learning and OSVI baselines on a subset of these tasks in Sec.~\ref{sec:r1}.

\subsection{Comparison with Baselines on Novel Tasks}\label{sec:r1}

\begin{figure}[!tp]
\centering
\includegraphics[width=\textwidth]{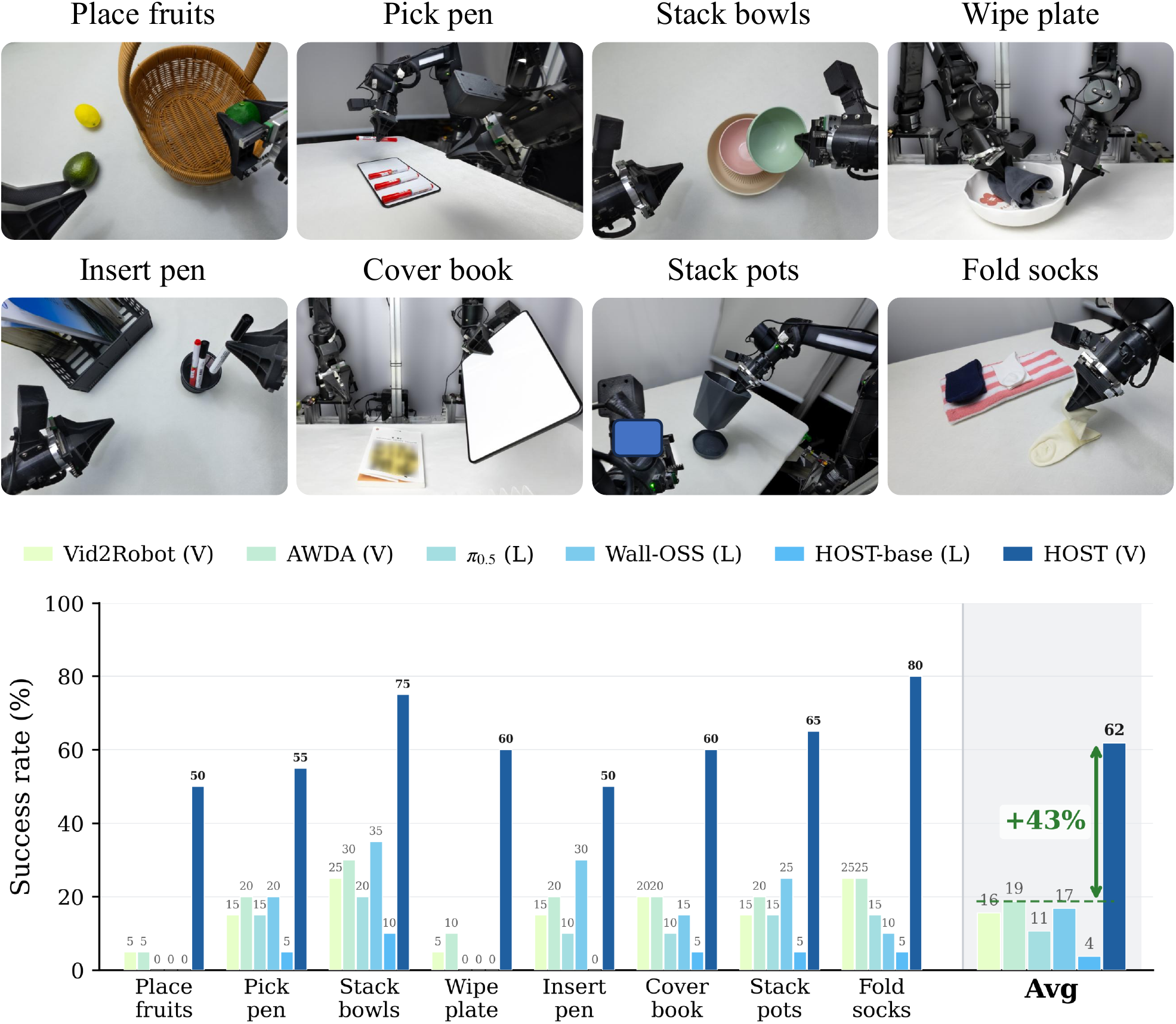}
\caption{\textbf{Success rates on novel manipulation tasks.} \method acquires each skill from a single human video without any parameter update across novel tasks whose manipulation skills are absent from the training data, each evaluated over 20 trials. \method is compared against OSVI baselines (Vid2Robot, AWDA) and language-conditioned imitation learning baselines ($\pi_{0.5}$, Wall-OSS, \method-base) evaluated zero-shot from the instruction, exceeding the strongest baseline average by 43\%, with the best result per task marked in \textbf{bold}.}
\label{fig:unseen_results}
\end{figure}

We compare \method against OSVI and language-conditioned imitation learning baselines on novel tasks whose manipulation skills are absent from the training data, with all methods evaluated without any parameter update. \method acquires the new skill from a single human video and attains the highest average success rate among all compared methods, as reported in Fig.~\ref{fig:unseen_results}. \method surpasses the strongest language-conditioned imitation learning baselines, $\pi_{0.5}$ and Wall-OSS, evaluated zero-shot from the instruction, by 45\%, despite both being pretrained on large-scale corpora. \method exceeds the strongest OSVI baseline by 43\% even though both methods receive the same single human video, indicating that its advantage comes not from access to a video demonstration but from how that video demonstration is used to drive execution, through the two mechanisms that distinguish \method, \couplemech{} and \stagemech{}, which together carry the demonstrated skill across the gap between human video and robot execution.

\subsection{Data and Time Efficiency of Skill Acquisition}\label{sec:data_eff}

\begin{figure}[!tp]
\centering
\includegraphics[width=\textwidth]{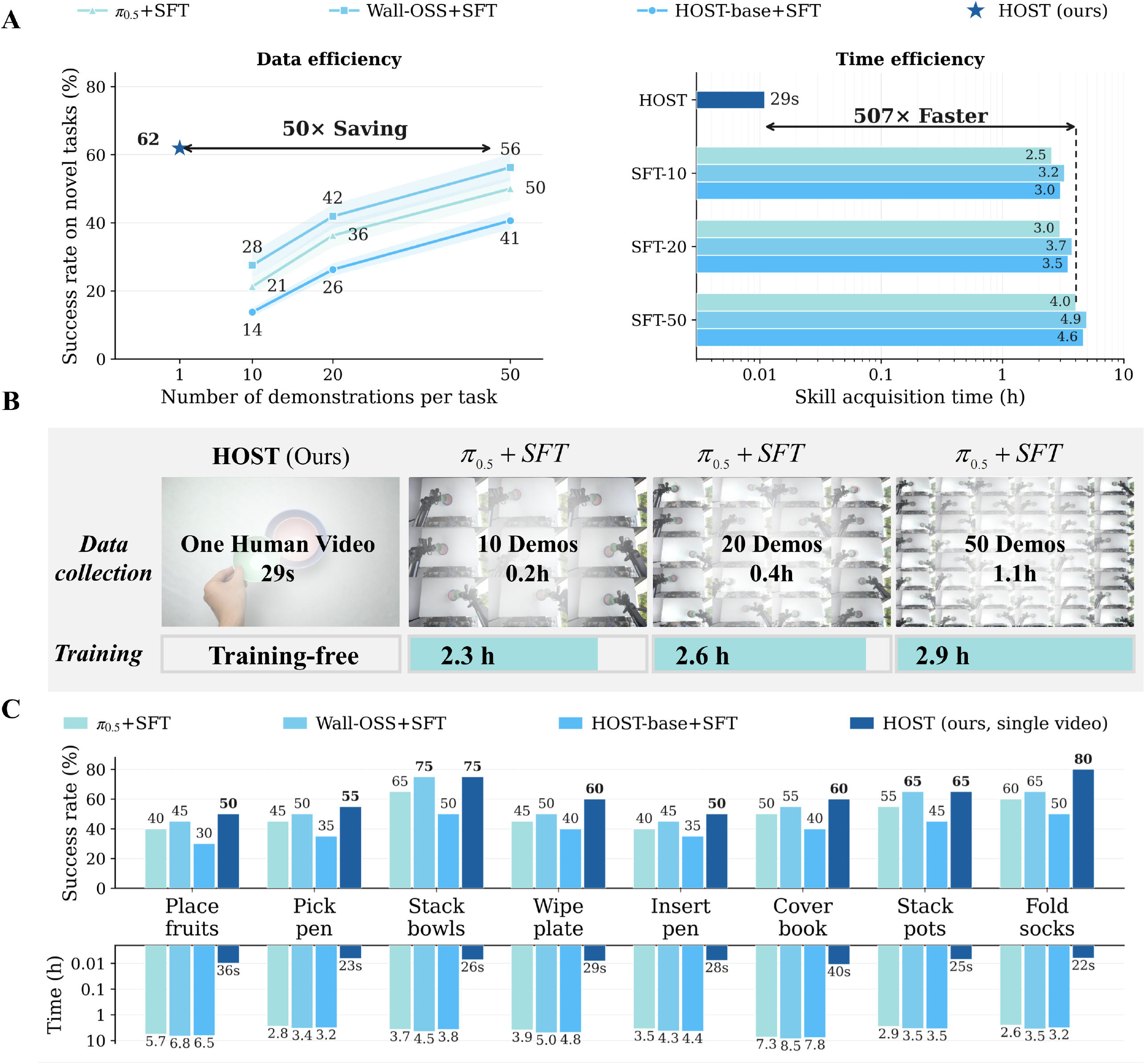}
\caption{\textbf{\method surpasses supervised fine-tuning in success rate, data efficiency and time efficiency.}
(\textbf{A})~Success rates on novel tasks and the time to acquire each skill for each SFT baseline at 10, 20 and 50 demonstrations per task, compared against the success rate and acquisition time of \method from a single human video.
(\textbf{B})~Qualitative illustration depicting the composition of the time to acquire each skill for $\pi_{0.5}$+SFT as a representative baseline, split between data collection and offline fine-tuning at 10, 20 and 50 demonstrations per task, compared against the time \method takes to acquire a skill from a single human video.
(\textbf{C})~Per-task success rates and the time to acquire each skill for each SFT baseline trained on 50 demonstrations per task, compared against \method from a single human video.}
\label{fig:data_time_efficiency}
\end{figure}

Sec.~\ref{sec:r1} establishes that \method exceeds the strongest baseline by 43\% on novel tasks. We further evaluate \method against imitation learning under supervised fine-tuning, the prevalent approach for teaching robots new skills, measuring the data and time efficiency of \method relative to it, as reported in Fig.~\ref{fig:data_time_efficiency}.

\textbf{Data efficiency.}
We fine-tuned each imitation learning baseline on 10, 20, and 50 teleoperated demonstrations per novel task using Low-Rank Adaptation (LoRA)~\cite{LoRA}. We then compared the resulting success rates with the performance of \method from a single human video, as presented in Fig.~\ref{fig:data_time_efficiency}A. Performance improved with the demonstration budget, but even the strongest fine-tuned baseline, Wall-OSS+SFT, reached only 56\% at 50 demonstrations per task, still 6\% below the 62\% \method reaches from a single video, while $\pi_{0.5}$+SFT and \method-base+SFT remained further below. From a single human video, \method therefore exceeds the performance of the strongest baseline at 50 demonstrations. The per-task comparison in Fig.~\ref{fig:data_time_efficiency}C confirms that this advantage holds across most of these tasks.

\textbf{Time efficiency.}
\method and the imitation learning baselines under supervised fine-tuning differ sharply in the time required to acquire each new skill. The cost of supervised fine-tuning is dominated by teleoperated collection of a demonstration corpus and offline fine-tuning, together taking approximately 4.0 to 4.9 hours per task at 50 demonstrations, as shown in Fig.~\ref{fig:data_time_efficiency}A and broken down in Fig.~\ref{fig:data_time_efficiency}B. \method, by contrast, reduces this collection to a single human video recording and removes offline fine-tuning entirely, making each skill available as soon as the video is recorded. \method compresses the time it takes to acquire each skill to approximately 29 seconds on average, 507 times faster than even the fastest SFT baseline, $\pi_{0.5}$+SFT, which requires 4.0 hours. The per-task comparison in Fig.~\ref{fig:data_time_efficiency}C confirms this gap holds consistently across individual tasks.

\subsection{Retention of Previously Mastered Skills}\label{sec:r_retention}

\begin{figure}[!tp]
\centering
\includegraphics[width=\textwidth]{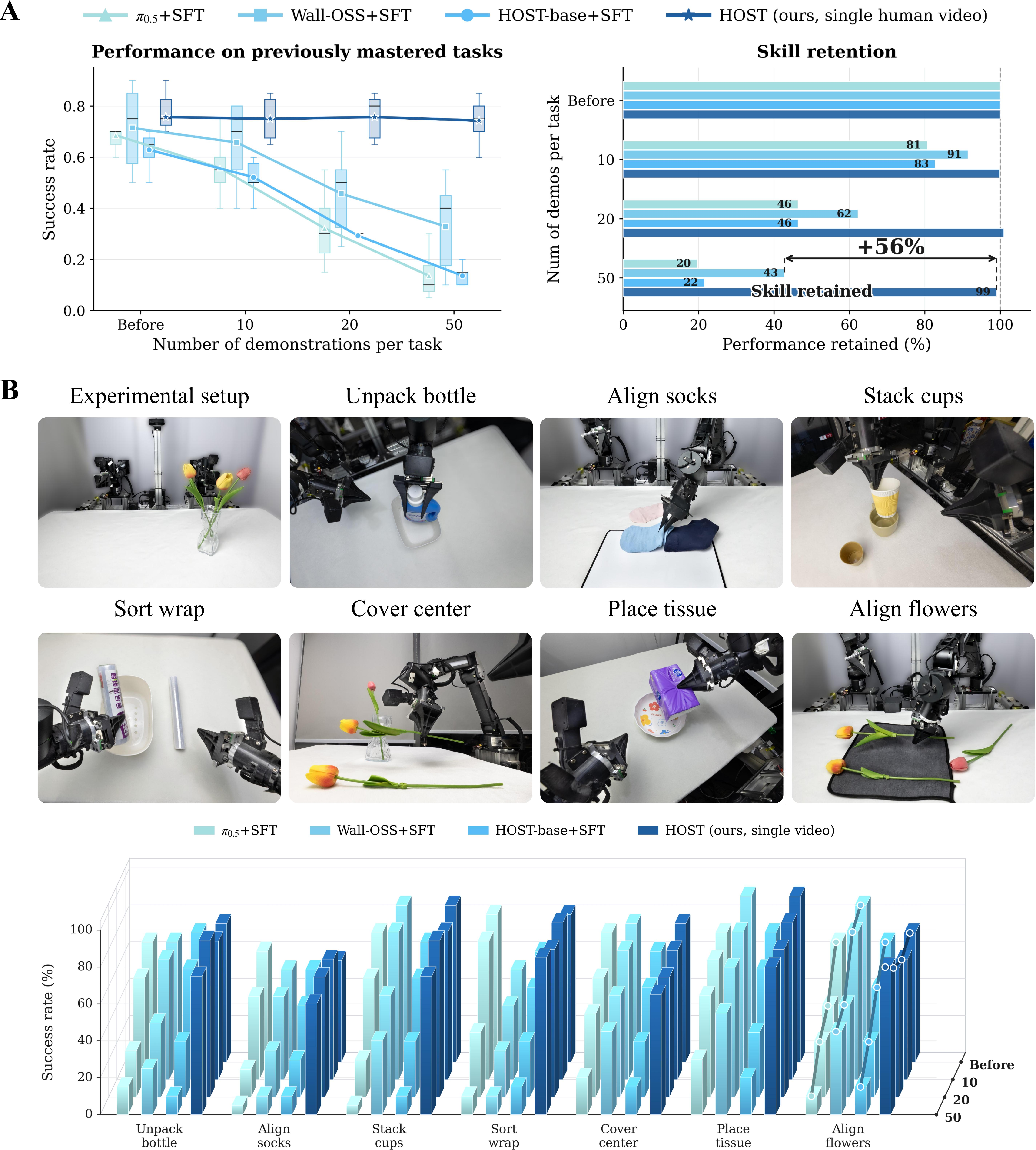}
\caption{\textbf{\method retains its performance on previously mastered tasks while supervised fine-tuning erodes it.}
(\textbf{A})~Success rate on previously mastered tasks for \method and each SFT baseline before and after acquiring a novel task, across demonstration budgets. Left, absolute success rate. Right, the percentage of performance retained relative to before acquisition.
(\textbf{B})~Per-task success rate on the seven previously mastered manipulation tasks, comparing \method against each SFT baseline before and after acquiring a novel task, across demonstration budgets.}
\label{fig:skill_retention}
\end{figure}

We further verify that \method acquires a new skill without eroding those already mastered, using the seven previously mastered tasks drawn from the training set and illustrated in Fig.~\ref{fig:skill_retention}B. For each method, we measured the success rate on these tasks before and after adapting to a novel task, as reported in Fig.~\ref{fig:skill_retention}A. Before adaptation, \method and the imitation learning baselines achieve comparable success rates on these tasks. \method attains the highest average from only a single video demonstration, as revealed in Fig.~\ref{fig:skill_retention}A and B. All three SFT baselines decline substantially from this comparable starting success rate after adaptation, with $\pi_{0.5}$+SFT retaining just \textcolor{\myemcolor}{20\%} of its original performance at 50 demonstrations per task, \method-base+SFT retaining \textcolor{\myemcolor}{22\%}, and Wall-OSS+SFT retaining \textcolor{\myemcolor}{43\%}, the highest retention among the three. \method, by contrast, acquires each new skill from the video without updating any parameters and retains that success rate afterward, with only a small residual difference arising from ordinary trial-to-trial variation. \method exceeds even the SFT baseline with the highest retention by \textcolor{\myemcolor}{56\%}. \method thus retains previously mastered skills where fine-tuning erodes them.

\subsection{Robustness under Deployment Perturbations}\label{sec:r5}

\begin{figure}[!tp]
\centering
\includegraphics[width=\textwidth]{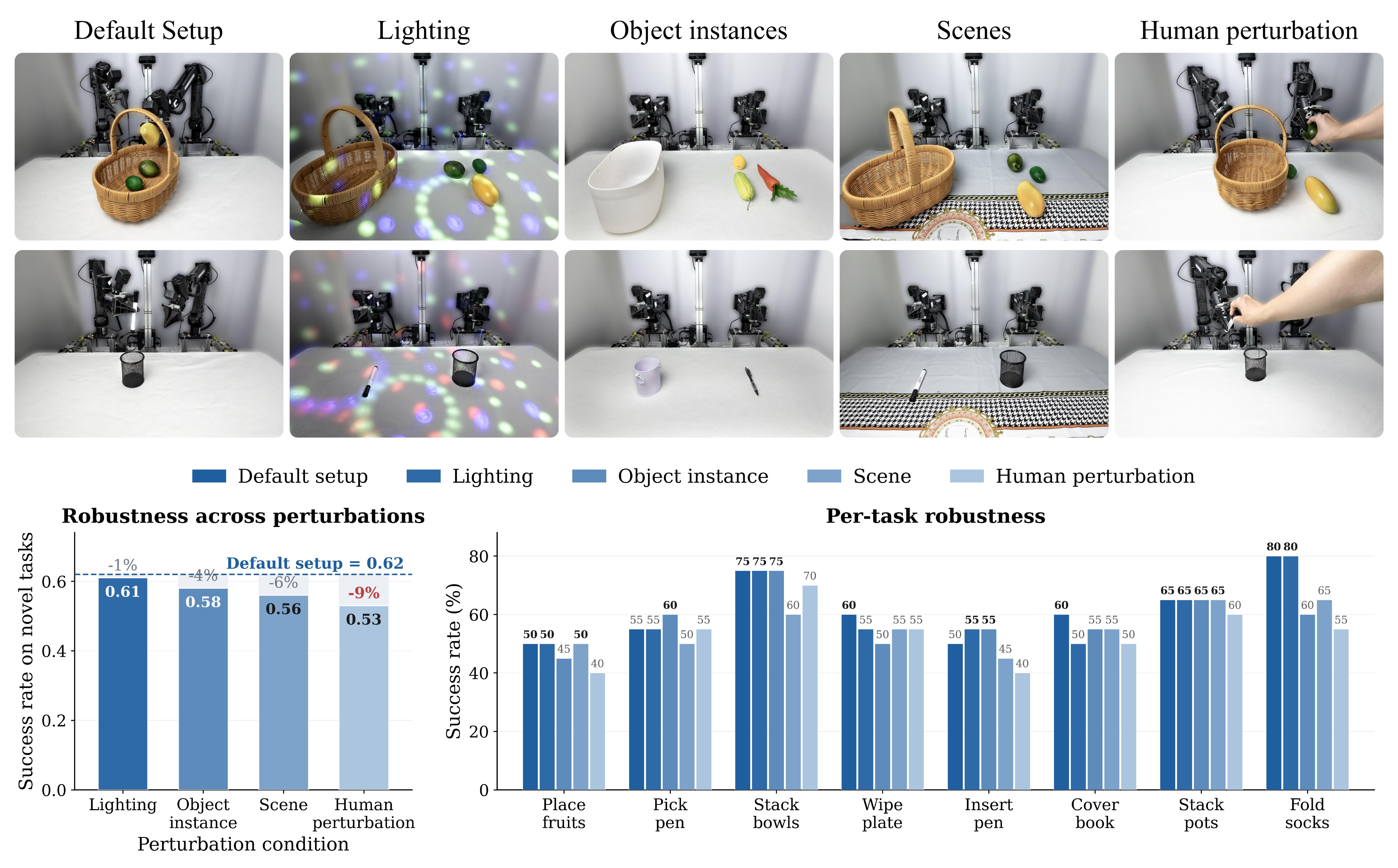}
\caption{\textbf{\method remains robust to deployment perturbations.} \method maintains strong performance under lighting variation, out-of-distribution objects, scene replacement, and human disturbance during execution, illustrated alongside the resulting average and per-task success rates on novel tasks under each condition.}
\label{fig:robustness}
\end{figure}

We evaluate the robustness of \method to four perturbations that increase the gap between the deployment environment and the video demonstration. Three perturbations alter the deployment environment, shifting lighting and color, substituting out-of-distribution objects, or replacing the scene. The fourth disrupts execution directly, with a person physically displacing an object while the robot is executing the task. \method retains the great majority of its default success rate of 62\% under all four conditions, as reported in Fig.~\ref{fig:robustness}. The performance loss increases with the severity of the perturbation, from 1\% under lighting shifts to 4\% under object substitution, 6\% under scene replacement, and 9\% under human disturbance during execution. Unlike the other three perturbations, the human disturbance occurs during execution itself, requiring \method to autonomously correct its course in response to an unexpected change. The per-task breakdown confirms that this resilience is broadly distributed. This robustness demonstrates that \method accurately localizes the current stage of robot execution within the video demonstration even when a human disturbance during execution shifts its task progress, and translates the upcoming progression of the video demonstration into corresponding robot behavior despite the state discrepancies introduced by changes in objects, scene, and lighting.

\subsection{Coupling Prediction Targets to the Demonstration}\label{sec:r2}

\begin{figure}[!tp]
\centering
\includegraphics[width=\textwidth]{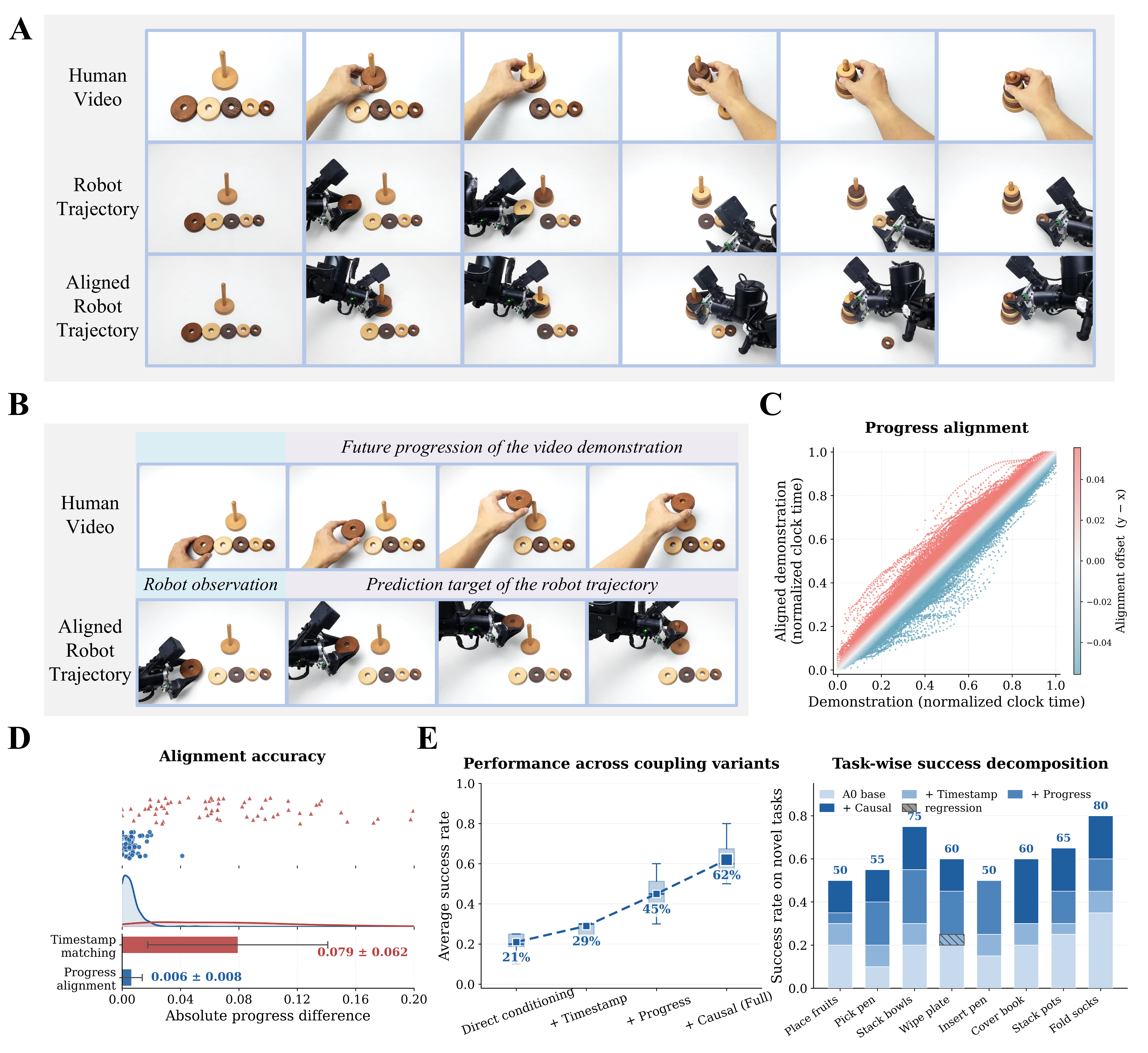}
\caption{\textbf{\Couplemech{} turns the video demonstration from a passive conditioning signal into the active driver of the robot's prediction.}
(\textbf{A})~Progress alignment on a shared task progress manifold recovers frame-level correspondence between the human video and the robot trajectory, whereas sampling based on clock time yields mismatched stages.
(\textbf{B})~Each robot prediction target is coupled to the upcoming progression of the demonstration.
(\textbf{C})~Aligned frame indices across paired training trajectories scatter around but not on the identity diagonal, confirming that correspondence based on clock time is insufficient and alignment is necessary.
(\textbf{D})~Progress alignment reduces the mean absolute progress error at human-annotated events from 0.079 under matching based on clock time to 0.006, an order-of-magnitude improvement.
(\textbf{E})~Each coupling ingredient contributes substantially to performance on novel tasks.}
\label{fig:alignment_ablation}
\end{figure}

\method couples each prediction target to the video demonstration, so that each target segment of the robot trajectory corresponds to the future evolution of the video demonstration. Fig.~\ref{fig:alignment_ablation} visualizes the alignment between the video demonstration and the robot trajectory, validates it against human annotations, and quantifies the contribution of each coupling ingredient to performance on novel tasks, the same held-out tasks evaluated in Fig.~\ref{fig:unseen_results}.

\textbf{Progress alignment and target coupling.} Coupling each prediction target to the video demonstration requires establishing frame-level correspondence between the robot trajectory and the video demonstration. One direct approach, correspondence based on clock time, matches frames by their relative position in time. \method instead aligns frames by task progress on a shared manifold, learned self-supervised without frame-level annotation. Fig.~\ref{fig:alignment_ablation}A indicates that this progress alignment recovers frame-level correspondence between the video demonstration and the robot trajectory, whereas correspondence based on clock time yields frames from mismatched task stages. Fig.~\ref{fig:alignment_ablation}B demonstrates that \method couples each prediction target to the upcoming progression of the video demonstration using this correspondence. Fig.~\ref{fig:alignment_ablation}C reveals that the frame indices recovered by progress alignment scatter around but not on the identity diagonal across paired training trajectories, confirming that correspondence based on clock time is structurally insufficient. Fig.~\ref{fig:alignment_ablation}D establishes that correspondence based on clock time yields a mean absolute progress difference of 0.079\,$\pm$\,0.062 from human-annotated reference points, compared with 0.006\,$\pm$\,0.008 for progress alignment. This order-of-magnitude improvement confirms the reliability of the self-supervised alignment module.

\textbf{Ablation of coupling variants.} Fig.~\ref{fig:alignment_ablation}E ablates each coupling ingredient in turn, isolating its contribution to performance on novel tasks. Conditioning the model on the entire video demonstration reaches 0.21. Restricting the model to a video demonstration window chosen by timestamp raises this to 0.29, a modest gain limited by the temporal asynchrony between the video demonstration and the robot trajectory, which often causes a timestamp-matched window to reflect a stage that the robot has not reached. Selecting the window by task progress on a shared manifold instead raises performance to 0.45, supplying the segment that matches the current execution stage. Coupling the prediction target to the future evolution of the video demonstration rather than to a fixed temporal offset along the execution trajectory further raises performance to 0.62, recovering the full model, with each ingredient in the coupling mechanism proving necessary for performance on novel tasks.

\subsection{Resolving Execution through Self-Grounded Prediction}\label{sec:r3}

\begin{figure}[!tp]
\centering
\includegraphics[width=\textwidth]{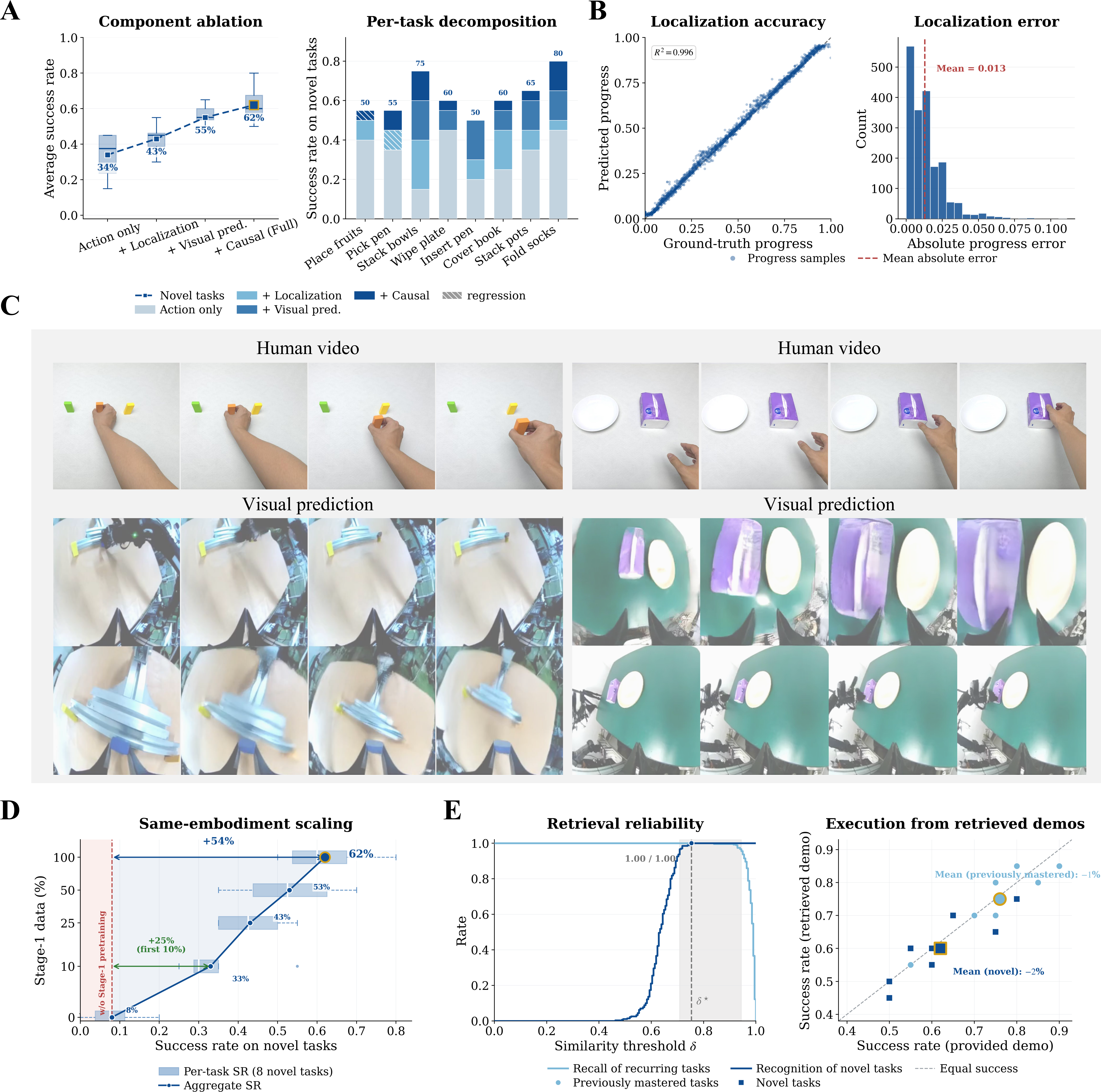}
\caption{\textbf{\Stagemech{} improves performance on novel tasks at each stage, same-embodiment pretraining scales skill acquisition from human video, and acquired skills persist across recurring tasks.}
(\textbf{A})~Ablation of \stagemech. Each added stage raises performance on novel tasks.
(\textbf{B, C})~Localization accuracy and visual prediction quality. Predicted progress positions closely track ground-truth values across all steps, and the model adapts the demonstrated behavior to the robot's own embodiment and deployment scene.
(\textbf{D})~Success on novel tasks increases monotonically with Stage~1 same-embodiment pretraining data volume, with Stage~2 adaptation data held fixed.
(\textbf{E})~Retrieval and execution reliability. Recall of recurring tasks and recognition of novel tasks are jointly high across a broad range of similarity thresholds, and execution from a retrieved demonstration matches the performance of a freshly provided one.}
\label{fig:ar_combined}
\end{figure}

\method resolves execution through \stagename{}, a causal cascade within a single autoregressive model that first localizes the robot's current stage within the video demonstration, then translates its upcoming progression into the robot's own future observations, and finally derives actions from these predicted observations. We quantify the contribution of each stage by introducing it in turn to a baseline that predicts action directly, as reported in Fig.~\ref{fig:ar_combined}A, with novel tasks drawn from Fig.~\ref{fig:unseen_results}. We then evaluate the accuracy of the localization stage and the quality of the predicted future observations that the cascade depends on, reported respectively in Fig.~\ref{fig:ar_combined}B and C.

\textbf{Ablation of \stagename{}.} Fig.~\ref{fig:ar_combined}A ablates each stage of \stagename{} in turn, isolating its contribution to performance on novel tasks. A baseline that maps the video demonstration directly to robot actions, without any intermediate stage, reaches 0.34. This baseline retains \couplemech{}, with \stagename{} therefore building further gains on top of target coupling. A localization stage raises success on novel tasks to 0.43, and a visual prediction stage running alongside action raises it further to 0.55. A causal cascade that derives actions from the predicted future observations raises performance to 0.62 and recovers the full model.

\textbf{Localization accuracy.}
We evaluate localization accuracy on the novel tasks, using ground-truth progress values provided by the alignment module. The predicted progress $\hat{p}_t$ closely tracks the ground-truth $p_t$ across all steps, as presented in Fig.~\ref{fig:ar_combined}B, with a mean absolute error of 0.013 in normalized progress. This error stays consistently small throughout, indicating that the localization stage tracks the progress of the robot within the video demonstration reliably.
  
\textbf{Visual prediction quality.}
Fig.~\ref{fig:ar_combined}C presents qualitative visual predictions on novel tasks. The predicted future observations follow the behavior demonstrated in the human video and translate it into the robot's own embodiment and deployment scene. This translation remains accurate across variations in object appearance and initial configuration, demonstrating the reliability of \stagename{} under variations in scene, viewpoint, and embodiment encountered at deployment.

\subsection{Effect of Same-Embodiment Pretraining}\label{sec:r4}

\method uses same-embodiment robot--robot pairs in Stage~1 to build the capacity to follow a demonstrated procedure, which Stage~2 then adapts to human video demonstrations using a smaller set of human--robot pairs. To quantify the contribution of this pretraining to skill acquisition from a single human video, we train variants on different fractions of the Stage~1 same-embodiment data while holding the Stage~2 adaptation data fixed, evaluating each variant on the same held-out novel tasks listed in Fig.~\ref{fig:unseen_results}. The 0\% condition, trained only on the limited Stage~2 human--robot pairs, performs poorly, confirming that these pairs alone are insufficient to build the capacity to follow a demonstrated procedure. As reported in Fig.~\ref{fig:ar_combined}D, success on novel tasks then increases monotonically as the volume of Stage~1 data grows. A larger volume of same-embodiment pretraining data exposes the model to a wider range of robot states and object configurations, strengthening its capacity to localize progress within a video demonstration, translate its upcoming progression into the robot's own future observations, and derive actions from those predicted observations. Stage~2 then adapts these capacities from robot videos to human video demonstrations. Same-embodiment pretraining therefore supplies the capacity needed for skill acquisition from a single human video, reducing reliance on scarce human--robot paired data.

\subsection{Persistence of Acquired Skills across Recurring Tasks}\label{sec:r_retrieval}

\method acquires each novel skill from a single human video. This video can then be stored and retrieved without repeated human involvement, enabling the acquired skill to persist across recurring tasks. \method thus accumulates these stored videos into a reusable store of previously acquired skills. We evaluate this persistence mechanism along three dimensions. We measure its accuracy in retrieving the correct video for a recurring task, its reliability in recognizing a genuinely novel task and requesting a new human video, and its execution reliability with a retrieved video relative to a freshly recorded one.

\textbf{Retrieval reliability.}
Each video demonstration is stored together with its task instruction and initial scene, enabling later encounters of the same or a similar task to retrieve it directly. Each requested task is posed as a query. Its instruction and current scene are compared with every stored entry through the weighted combination of instruction and scene similarity defined in Eq.~\ref{eq:retrieval_score}. The entry with the highest score above a threshold $\delta$ is retrieved. \method operates at a single fixed threshold $\delta^\star$ across all tasks, retrieving the correct stored video for a recurring task and recognizing a genuinely novel task, requesting a new human video instead, as shown in Fig.~\ref{fig:ar_combined}E. This threshold balances two error modes, a value set too low admits false matches that reuse an unrelated stored video, whereas a value set too high rejects valid matches and triggers unnecessary requests for a new human video. Retrieval accuracy and recognition of novel tasks remain high across a wide band of thresholds separating these two failure modes, and $\delta^\star$ is chosen well within that band. This wide margin keeps \method reliable regardless of the exact choice of $\delta$.

\textbf{Execution reliability from retrieved videos.}
Recurring tasks proceed without repeated human involvement, which requires retrieval to both locate the correct video and support execution as reliable as a freshly recorded one. \method acquires and executes each skill without any parameter update, and retrieval therefore substitutes only the source of the video demonstration, with the same frozen model processing it exactly as it would a freshly recorded video. Fig.~\ref{fig:ar_combined}E compares execution driven by a retrieved video with execution driven by a freshly recorded video on both previously mastered and novel tasks. The corresponding success rates match the \method averages reported in Figs.~\ref{fig:skill_retention} and~\ref{fig:unseen_results}. The two settings achieve comparable performance with only a small residual gap.

\section{Discussion}

The ability to acquire new skills rapidly and effortlessly while retaining those already mastered is essential for robots operating in human environments. However, current methods still acquire each new skill through a cumbersome training-time loop that is costly, slow and self-defeating. Collecting task-specific demonstrations consumes expert labor and scarce robot time, offline fine-tuning extends over hours before each new skill becomes available, and updating the shared policy parameters erodes skills already mastered. We therefore propose \method, which moves novel skill acquisition from this loop to inference. From a single human video, \method acquires each new skill and executes it without any parameter update.

Realizing this paradigm, however, confronts a fundamental problem rooted in the structural mismatch between video demonstration and execution. Temporal asynchrony between the video demonstration and the robot execution decouples the prediction target from the video demonstration. State discrepancies in embodiment, viewpoint and appearance further hinder translating the observed behavior into action. Addressing this mismatch calls for a new formulation of how a robot acquires a skill from a demonstration. The central idea of \method is to treat skill acquisition as a process driven by the video demonstration and resolved through \stagename{}, built on two principles. First, the video demonstration actively determines the robot's prediction target. \method couples this target to the upcoming progression of the video demonstration on a \manifold{}, turning the video into the active driver of the robot's prediction. Second, \method resolves execution through a causal cascade of \stagename{}. This cascade first localizes where the robot stands within the demonstrated procedure, then translates its upcoming progression into the robot's own future observations, and finally derives actions from these predicted observations. \method thereby enables the robot to actively follow the demonstrated procedure and adapt it to the robot's own embodiment.

\method acquires manipulation skills in seconds from a single human video, while retaining those already mastered. \method reaches 62\% average success on novel tasks without any parameter update, acquiring each skill in approximately 29 seconds. \method exceeds the strongest one-shot visual imitation baseline by 43\% even though that baseline receives the same single human video, and the strongest zero-shot imitation learning baseline by 45\%. Even compared with the strongest baseline fine-tuned on 50 robot demonstrations per task, \method achieves superior performance while using 50 times fewer demonstrations and acquiring each new skill 507 times faster. This performance does not come at the cost of previously mastered skills. \method retains these skills, whereas even the strongest fine-tuned baseline retains only \textcolor{\myemcolor}{43\%} of its original performance. A human video can also be stored and retrieved to support recurring tasks without repeated human involvement.

\method may also point toward a new possibility for skill acquisition in embodied agents. A robot could autonomously and continually learn new skills by observing the people around it. For example, a household robot could observe the people it lives with, while a factory robot could observe nearby workers. Realizing this requires two capabilities, learning new skills directly from human video and continuing to acquire them over time. \method, built on \couplemech{} and \stagemech{}, operates directly on human video and enables the robot to actively follow the observed activity and adapt it to its own embodiment without any parameter update. This allows a robot to learn new skills directly from human video at low cost. Each video keeps the acquired skill in external context rather than encoding it in the policy weights. This video can then be stored and retrieved without repeated human involvement, enabling the acquired skill to persist across recurring tasks. A robot can thus accumulate these stored demonstrations without eroding previously acquired skills, continuing to acquire new skills over time.

Despite its contributions, this work has several limitations that warrant further investigation. First, our evaluations are conducted on a single bimanual manipulation platform, and \method remains untested on robots that differ substantially in embodiment. Training on human--robot paired data from a wider range of robots would extend \method to these platforms. Second, video demonstrations capture the manipulation procedure without conveying the contact forces relevant to fine-grained manipulation~\cite{Fazeli2019SeeFeelAct}. Capturing these contact forces, for instance with wearable sensors such as tactile gloves, is a promising direction for future iterations. Finally, \method retrieves stored demonstrations by combining instruction and scene similarities between each stored demonstration and the requested task. As the memory grows, this similarity may not reliably separate tasks whose instructions and scenes differ only subtly, and capturing these finer distinctions is a direction for future work.

\section{Methods}

\subsection{Problem Formulation and Method Overview}\label{sec:prob_formulation}

\textbf{Problem formulation.} Let $\mathcal{M} = \mathcal{M}_{\mathrm{train}} \cup \mathcal{M}_{\mathrm{test}}$ denote a set of manipulation tasks partitioned into disjoint training and testing subsets. Each task $m \in \mathcal{M}_{\mathrm{train}}$ is associated with a language instruction $\ell$, a set of video demonstrations $\{\tau^d_i\}_{i=1}^{N^d}$, and a set of robot trajectories $\{\tau^r_j\}_{j=1}^{N^r}$ that accomplish the same manipulation goal under potentially different embodiments and scene configurations. A video demonstration $\tau^d_i = \{o^d_t\}_{t=1}^{T_i}$ records a demonstrator performing the task as a sequence of $T_i$ visual observations, with $o^d_t \in \mathcal{O}_d$. A robot trajectory $\tau^r_j = \{(o^r_t, s^r_t, a_t)\}_{t=1}^{N_j}$ records the robot carrying out the same task and comprises observations $o^r_t \in \mathcal{O}_r$, proprioceptive states $s^r_t \in \mathcal{S}$, and executed actions $a_t \in \mathcal{A}$. Although a video demonstration and a trajectory realize the same manipulation goal, they differ in length, with $N_j$ generally differing from $T_i$, and may further differ in viewpoint, embodiment, execution dynamics, scene layout, and object instance. The model, trained on $\mathcal{M}_{\mathrm{train}}$, is evaluated on a disjoint test set $\mathcal{M}_{\mathrm{test}}$, where each task provides only a single video demonstration. The objective is to learn a policy $\pi_\theta(a_t \mid \tau^d, \ell, o^r_{t\text{-}K:t}, s^r_t)$ that, given a video demonstration, a language instruction, recent observations and the current proprioceptive state of the robot, generalizes to novel tasks without parameter updates, thereby enabling inference-time skill acquisition while preserving previously learned capabilities.

\textbf{Method overview.} \method is built on a mechanistic view of skill acquisition from a single video demonstration. Such a video demonstration does not directly specify the actions for robot execution, and a structural mismatch exists between the video demonstration and robot execution. The video demonstration and robot execution unfold asynchronously, decoupling the prediction target from the video demonstration and relegating it to a passive context. The video demonstration and the robot trajectory also differ in embodiment, viewpoint and appearance, and these discrepancies hinder the direct translation of the observed behaviour into robot action.

As illustrated in Fig.~\ref{fig:method_overview}, \method comprises two core mechanisms, \couplemech{} and \stagemech{}, that together resolve the structural mismatch between the video demonstration and robot execution. First, it couples the prediction target at each robot timestep to the upcoming progression of the video demonstration (Sec.~\ref{sec:alignment}), through a monotonic frame-level correspondence that maps the video demonstration and the robot trajectory onto a \manifold, learned self-supervised without frame-level annotation. Second, it predicts the target through a cascade of \stagename{} (Sec.~\ref{sec:wam}). Given a video demonstration, recent robot observations $o^r_{t\text{-}K:t}$ and state $s^r_t$, and an optional language instruction $\ell$, it first localizes the robot's current stage within the video demonstration, then translates its upcoming progression into the robot's own future observations, and finally derives robot actions. It is pretrained on large-scale same-embodiment paired data and adapted to human-to-robot scenarios with a small set of human--robot pairs.

After training, \method is able to acquire novel skills from a single human video demonstration at inference time. Given a single human video of an unseen task, the video drives the same coupled, self-grounded prediction, and the robot acts without any parameter update. A new skill therefore enters through the video at inference time, relocating skill acquisition from the training-time loop to inference.

\subsection{Coupling Prediction Targets to the Demonstration}\label{sec:alignment}

A video demonstration and a robot execution of the same task are not temporally synchronized, unfolding as two independent processes at different speeds. This decouples the prediction target, tied to progress along the robot trajectory, from the video demonstration content, relegating the video demonstration to a passive context and hindering the model from acquiring novel skills from the video demonstration at inference time. We address this by \couplemech{}. The prediction target at each robot timestep is redefined as a dynamic segment of the robot's own future trajectory, which step-wise corresponds to the upcoming evolution of the video demonstration. These coupled targets are constructed from a monotonic frame-level correspondence recovered by an alignment module, trained self-supervised on same-task trajectory pairs from $\mathcal{M}_{\mathrm{train}}$ with Smooth Dynamic Time Warping~\cite{hadji2021} and temporal cycle-consistency~\cite{dwibedi2019tcc}, without frame-level annotation.

\textbf{Task progress alignment.} Given a video demonstration $\tau^d = \{o^d_t\}_{t=1}^{T}$ and a robot trajectory $\tau^r = \{o^r_t\}_{t=1}^{N}$ of the same task, a shared frame embedding model $f_\phi$, built on Qwen3-VL-Embedding-8B~\cite{qwen3vl} and trained end-to-end during alignment training, encodes each frame into a $d_{\mathrm{emb}}$-dimensional embedding. For each timestep of the trajectory, multi-view observations from all cameras are organized as an image sequence and fed to the model, with an [EOS] token appended at the end. The per-timestep representation $h_t \in \mathbb{R}^D$ is derived from the hidden state of the last layer corresponding to this [EOS] token. A linear projection followed by $\ell_2$ normalization maps $h_t$ into a $d_{\mathrm{emb}}$-dimensional embedding $\mathbf{e}_t$:
{\small\begin{equation}\label{eq:embedding}
  h_t = f_\phi(o_t), \quad
  \mathbf{e}_t = \frac{P h_t}{\|P h_t\|_2},
\end{equation}}
where $P \in \mathbb{R}^{d_{\mathrm{emb}} \times D}$ is a learnable projection matrix. Applying the same model to the video demonstration and the robot trajectory yields embedding sequences $\mathbf{d} = \{\mathbf{d}_1, \ldots, \mathbf{d}_T\}$ and $\mathbf{r} = \{\mathbf{r}_1, \ldots, \mathbf{r}_N\}$ of lengths $T$ and $N$, respectively.

A pairwise similarity matrix $S \in \mathbb{R}^{T \times N}$ is computed from the negative squared L2 distance between embeddings, scaled by a temperature parameter $\kappa$. Column-wise log-softmax normalization~\cite{hadji2021} then converts $S$ into a pairwise cost matrix $C$, controlled by a column-normalization temperature $\gamma_f$:
{\small\begin{equation}\label{eq:sim_cost}
  S_{ij} = -\|\mathbf{d}_i - \mathbf{r}_j\|^2 / \kappa, \quad c(i,j) = -\log \frac{\exp(S_{ij} / \gamma_f)}{\sum_{k=1}^{T} \exp(S_{kj} / \gamma_f)}.
\end{equation}}
Smooth Dynamic Time Warping~\cite{hadji2021} then recovers monotonic soft matching probabilities from the cost matrix through a forward--backward dynamic programming procedure. The forward pass computes a cumulative cost table $R$, in which $R(i,j)$ is the cost of the optimal monotonic path from $(1,1)$ to $(i,j)$:
{\small\begin{equation}\label{eq:sdtw_forward}
  R(i,j) = c(i,j) + \mathrm{smoothMin}\!\big(R(i\!-\!1,j\!-\!1),\; R(i\!-\!1,j),\; R(i,j\!-\!1);\; \gamma\big),
\end{equation}}
where $\mathrm{smoothMin}$ is a differentiable approximation to the minimum operator, parameterized by a smoothness constant $\gamma > 0$ that reduces to the hard minimum as $\gamma \to 0$:
{\small\begin{equation}\label{eq:smoothmin}
  \mathrm{smoothMin}(\mathbf{a};\gamma) = \frac{\sum_k a_k \exp(-a_k/\gamma)}{\sum_l \exp(-a_l/\gamma)}.
\end{equation}}
The backward pass computes $E$ analogously, in which $E(i,j)$ is the cost of the optimal monotonic path from $(i,j)$ to $(T,N)$. The soft matching probabilities are obtained by combining $R$ and $E$ and normalizing each row:
{\small\begin{equation}\label{eq:sdtw_beta}
  \beta^{d \to r}_{ij} = \frac{\exp\!\big({-\gamma^{-1}(R(i,j) + E(i,j))}\big)}{\sum_{k=1}^{N} \exp\!\big({-\gamma^{-1}(R(i,k) + E(i,k))}\big)},
\end{equation}}
where $\beta^{d \to r}_{ij}$ represents the probability that timestep $i$ of $\mathbf{d}$ corresponds to timestep $j$ of $\mathbf{r}$. The monotonic transition structure in Eq.~\ref{eq:sdtw_forward}, which only permits advancing along $i$, $j$, or both, constrains $\beta^{d \to r}$ to concentrate on temporally ordered correspondences. The reverse matching probabilities $\beta^{r \to d} \in \mathbb{R}^{N \times T}$ are computed analogously from $S^\top$.

\textbf{\Couplemech.} The monotonic transition structure of Smooth DTW encourages the learned matching distributions to concentrate along temporally ordered paths. This structure allows direct extraction of bidirectional frame correspondences through rowwise maximum probability assignment.
{\small\begin{equation}\label{eq:alignment_map}
  \pi^{r \to d}(t) = \arg\max_j\, \beta^{r \to d}_{tj}, \quad
  \pi^{d \to r}(t) = \arg\max_j\, \beta^{d \to r}_{tj}.
\end{equation}}
For a robot at timestep $t$, the mapping $\pi^{r \to d}$ locates the video demonstration frame corresponding to the current execution stage. The $H$ video demonstration frames following this position define the upcoming video demonstration segment. Mapping each of these frames back to the robot trajectory via $\pi^{d \to r}$ yields $H$ corresponding robot timesteps, and the robot trajectory segment $\mathcal{T}_t$ at these timesteps constitutes the prediction target:
{\small\begin{equation}\label{eq:pred_target}
  t_i = \pi^{d \to r}\!\big(\pi^{r \to d}(t) + i\big), \quad i = 0, 1, \ldots, H\!-\!1, \quad
  \mathcal{T}_t = \{\tau^r_{t_0}, \ldots, \tau^r_{t_{H-1}}\}.
\end{equation}}
Since the mappings are monotonic, the target comprises an ordered sequence of aligned robot timesteps whose temporal extent is determined by the alignment.

This formulation strictly binds the prediction target $\mathcal{T}_t$ to the video demonstration, with $\mathcal{T}_t$ directly defined by the video demonstration segment $\{o^d_{\pi^{r \to d}(t)}, \ldots, o^d_{\pi^{r \to d}(t)+H-1}\}$. To reduce encoding cost and enable the model to attend to the video demonstration content most relevant to the current robot state, a sliding window $\mathcal{W}$ of $L \geq H$ video demonstration frames centered around the coupled segment is randomly selected and fed to the model in place of the full video demonstration $\tau^d$. The current progress of the robot within $\mathcal{W}$ is recorded as a scalar localization label $p_t$, supervising the model to localize its current task stage within the video demonstration window and adaptively extract the relevant segment, which further enables autonomous window advancement at inference time. Each training sample is formally defined as a tuple $(\mathcal{W},\, o^r_{t-K:t},\, s^r_t,\, \ell,\, p_t,\, \mathcal{T}_t)$, where $\mathcal{T}_t$ comprises a future observation component $\mathcal{T}_t^o = \{o^r_{t_0}, \ldots, o^r_{t_{H-1}}\}$ and an action component $\mathcal{T}_t^a = \{a_{t_0}, \ldots, a_{t_{H-1}}\}$.

\textbf{Self-supervised alignment objective.} To optimize the embedding network in a self-supervised manner and obtain accurate matching probabilities $\beta$, two complementary losses are applied. We describe the $d \to r$ direction below. The reverse direction follows by symmetry.

A temporal cycle-consistency loss~\cite{dwibedi2019tcc} enforces frame-level correspondence by constraining each frame to cycle back to itself, matched to its nearest neighbor in the other sequence and then matched back. The loss penalizes the temporal deviation between the original frame and the cycled-back frame. For each video demonstration frame $\mathbf{d}_i$, the soft nearest neighbor $\tilde{\mathbf{r}}_i$ in $\mathbf{r}$ and the cycle-back distribution $\hat{\beta}_i$ over $\mathbf{d}$ are computed as:
{\small\begin{equation}\label{eq:soft_nn}
  \tilde{\mathbf{r}}_i = \sum_{j=1}^{N} \beta^{d \to r}_{ij}\, \mathbf{r}_j, \quad
  \hat{\beta}_{ik} = \frac{\exp(\tilde{\mathbf{r}}_i^\top \mathbf{d}_k / \kappa)}{\sum_{l=1}^{T} \exp(\tilde{\mathbf{r}}_i^\top \mathbf{d}_l / \kappa)}.
\end{equation}}
From $\hat{\beta}_i$, the cycle-back prediction $\mu_i$ and its variance $\nu_i^2$ are derived, and the loss penalizes deviations of $\mu_i$ from the original index $i$:
{\small\begin{equation}\label{eq:tcc_loss}
  \mu_i = \sum_{k} \hat{\beta}_{ik}\, k, \quad
  \nu_i^2 = \sum_{k} \hat{\beta}_{ik}\, (k - \mu_i)^2, \quad
  \mathcal{L}_{\mathrm{TCC}} = \frac{1}{T} \sum_{i=1}^{T} \left[\frac{(i - \mu_i)^2}{\nu_i^2} + \lambda \log \nu_i\right].
\end{equation}}

To further encourage globally coherent alignments beyond per-frame consistency, a DTW loss minimizes the total alignment cost between the two sequences, computed as the forward table value at the terminal cell normalized by sequence length:
{\small\begin{equation}\label{eq:d2tw_loss}
  \mathcal{L}_{\mathrm{DTW}} = \frac{R(T,N)}{T + N}.
\end{equation}}
The two losses are symmetrized over both directions, and the total alignment loss is $\mathcal{L}_{\mathrm{align}} = \mathcal{L}_{\mathrm{TCC}} + \lambda_{\mathrm{DTW}}\, \mathcal{L}_{\mathrm{DTW}}$.

\subsection{Resolving Execution through Self-Grounded Prediction}\label{sec:wam}

Beyond temporal alignment, the video demonstrations and the robot trajectories differ in embodiment, viewpoint and appearance. Directly deriving robot actions from the video demonstration forces the model to resolve intractable cross-domain ambiguities, hindering translation into precise action. Inspired by the causal cascade of human observational learning~\cite{schacter2007prospective,meltzoff1988infant,gershman2017reinforcement,jeannerod2001neural,wolpert2003unifying,rizzolatti2001neurophysiological,buccino2004neural}, we propose \stagemech{}. We localize the robot's current progress within the video demonstration, translate its upcoming progression into the robot's own future observations conditioned on the localized segment, and derive robot actions from these future observations. We model this causal cascade within a single autoregressive diffusion model, as illustrated in Fig.~\ref{fig:method_overview}B. The generation targets are arranged in a causal token sequence in which each output is conditioned on all preceding ones and jointly optimized in a single forward pass.

\textbf{Input representation.} All visual elements, including the video demonstration window $\mathcal{W}$, recent robot observations $o^r_{t\text{-}K:t}$, and the future observations $\mathcal{T}_t^o$, are encoded into a shared latent space by the Wan VAE~\cite{wan2025}, yielding video demonstration tokens $z^d$, robot observation tokens $z^o$, and target observation tokens $y^o$. The localization value $p_t$ and the action sequence $\mathcal{T}_t^a$ are projected to the hidden dimension of the model through separate learnable linear layers, producing a localization token $y^p = E_p(p_t)$ and action tokens $y^a$. To implement autoregressive generation via flow matching, each generation target is represented as a token pair, one noisy and one clean, where the noisy token serves as the denoising input and the clean token conditions subsequent targets:
{\small\begin{equation}\label{eq:noise}
  y^p_{\sigma} \!=\! (1 \!-\! \sigma)\,y^p + \sigma\,\epsilon_p, \quad
  y^o_{\sigma} \!=\! (1 \!-\! \sigma)\,y^o + \sigma\,\epsilon_o, \quad
  y^a_{\sigma} \!=\! (1 \!-\! \sigma)\,y^a + \sigma\,\epsilon_a.
\end{equation}}
The complete token sequence places the context tokens first, followed by the target pairs in causal order:
{\small\begin{equation}\label{eq:sequence}
  [\,z^d \;\|\; z^o \;\|\; y^p_{\sigma},\, y^p \;\|\; y^o_{\sigma},\, y^o \;\|\; y^a_{\sigma}\,],
\end{equation}}

\textbf{Architecture.} The model adopts a dual-expert Mixture-of-Transformer~\cite{MoT} architecture built on the Wan video diffusion backbone~\cite{wan2025}. A \videoexpert, initialized from Wan2.2~\cite{wan2025}, handles localization and future observation prediction. An \actionexpert with the same depth but reduced hidden dimension handles action generation, a reduction afforded by the lower distributional complexity of actions relative to observations. The two experts share self-attention by concatenating their respective query, key, and value projections at each layer, with $\mathbf{Q}_\mathrm{ve}$, $\mathbf{K}_\mathrm{ve}$, $\mathbf{V}_\mathrm{ve}$ for the \videoexpert and $\mathbf{Q}_\mathrm{ae}$, $\mathbf{K}_\mathrm{ae}$, $\mathbf{V}_\mathrm{ae}$ for the \actionexpert:
{\small\begin{equation}\label{eq:mot_attn}
  \mathbf{Q} = [\mathbf{Q}_\mathrm{ve} \,|\, \mathbf{Q}_\mathrm{ae}], \quad \mathbf{K} = [\mathbf{K}_\mathrm{ve} \,|\, \mathbf{K}_\mathrm{ae}], \quad \mathbf{V} = [\mathbf{V}_\mathrm{ve} \,|\, \mathbf{V}_\mathrm{ae}],
\end{equation}}
The joint attention output is then split and processed by expert-specific feed-forward networks. Both experts additionally receive a cross-attention conditioning sequence $c$ at every layer, consisting of the language instruction encoded by the \textenc{} text encoder~\cite{umt5} and the proprioceptive state $s^r_t$ projected via a learnable matrix $M_s$:
{\small\begin{equation}\label{eq:context}
  c = [\mathrm{UMT5}(\ell) \,|\, M_s\, s^r_t].
\end{equation}}

\textbf{Autoregressive causal structure.} The causal order among the localization token, the future observation tokens, and the action tokens is enforced through the self-attention mask. Each noisy target token attends to the video demonstration tokens $z^d$, the robot observation tokens $z^o$, and the clean tokens of all preceding targets, while attention to concurrent and subsequent targets is blocked. During training, the clean tokens are set to ground-truth representations, and gradients through the clean tokens are detached so that gradients from each target loss do not propagate through preceding clean targets. The localization, observation, and action tokens are therefore optimized jointly in a single forward pass. To bridge the gap between training and inference, the ground-truth localization value and future observations serving as clean tokens are augmented with small-magnitude Gaussian noise during training.

\textbf{Training objective.} The model predicts a velocity for the noisy token of each target following the causal order established above, denoted $\hat{v}_p$, $\hat{v}_o$ and $\hat{v}_a$ for the localization, observation and action targets. Each is trained with a flow matching objective~\cite{lipman2023flow} that compares the predicted velocity against the target:
{\small\begin{align}\label{eq:flow_matching}
  \mathcal{L}_{\mathrm{loc}} &= \left\| \hat{v}_p - (\epsilon_p - y^p) \right\|^2, \notag\\
  \mathcal{L}_{\mathrm{obs}} &= \left\| \hat{v}_o - (\epsilon_o - y^o) \right\|^2, \\
  \mathcal{L}_{\mathrm{act}} &= \left\| \hat{v}_a - (\epsilon_a - y^a) \right\|^2. \notag
\end{align}}
The total training loss is formulated as:
{\small\begin{equation}\label{eq:total_loss}
  \mathcal{L} = \lambda_p \mathcal{L}_{\mathrm{loc}} + \lambda_o \mathcal{L}_{\mathrm{obs}} + \lambda_a \mathcal{L}_{\mathrm{act}}.
\end{equation}}

\subsection{Training and Inference}
\label{sec:training_inference}

\textbf{Two-stage training.}
Human--robot paired data would directly supervise the re-expression of a demonstrated procedure across embodiment, but such data remain scarce, impeding the development of reliable skill acquisition from a single human video. \method instead develops the capacity to follow a demonstrated procedure first, then adapts it to human video demonstrations through a two-stage training protocol that reduces reliance on this scarce supervision. Stage~1 develops this capacity from readily available same-embodiment robot--robot pairs, and Stage~2 adapts it to human video demonstrations with a smaller set of human--robot pairs.

In Stage~1, \method is pretrained on large-scale same-embodiment robot--robot pairs. Different executions of the same task provide natural supervision for prediction conditioned on a video demonstration, with one trajectory serving as the video demonstration and another providing the robot's own future observations and actions. This stage teaches the model to localize progress, predict future observations and generate actions conditioned on a demonstrated procedure, without requiring cross-embodiment transfer. The video demonstrations used in this stage are exclusively robotic.

In Stage~2, the pretrained model is adapted with a smaller collection of human--robot paired demonstrations. The prediction targets remain drawn from the robot's own future trajectory, whereas the conditioning video demonstration is drawn from human video. This stage adapts the mechanism learned from robot--robot pairs to the harder setting in which task structure must be recovered across discrepancies that extend beyond viewpoint and appearance to a further difference in embodiment. Both stages optimize the same objective in Eq.~\ref{eq:total_loss}, allowing the model to reuse the same \stagename{} across same-embodiment and cross-embodiment settings.

The alignment module described in Sec.~\ref{sec:alignment} is pretrained on paired same-task trajectories spanning both same-embodiment and cross-embodiment settings. Once trained, it is used to construct training samples for the policy model, with prediction targets coupled to the video demonstration.

\textbf{Inference-time skill acquisition.}
After training, \method acquires each new skill at inference time from a video, which serves as an external specification of the task and drives execution without any per-task training. Given a video demonstration $\tau^d$ and a language instruction $\ell$, the model initializes the video demonstration window at frame index $w_0=0$ and repeatedly performs the same causal inference loop, localizing the robot's current progress within the video demonstration, predicting the robot's own future observations from the upcoming progression of the video demonstration, and generating an action chunk to realize those observations. The video demonstration window is advanced according to the predicted progress, allowing the robot to follow the demonstrated task structure.

At step $t$, the model receives the current video demonstration window $\mathcal{W}_t$, recent robot observations $o^r_{t-K:t}$, proprioceptive state $s^r_t$ and language instruction $\ell$. The localization token $y^p$, the future observation tokens $y^o$ and the action tokens $y^a$ are generated sequentially from noise, with each completed output conditioning subsequent targets:
{\small
\begin{equation}
\label{eq:ar_inference}
  \hat{y}_i
  =
  \mathrm{Denoise}(z^d,z^o,c,\hat{y}_{<i}),
  \qquad
  i\in\{p,o,a\},
\end{equation}
}
where $\hat{y}_{<i}$ denotes the preceding generated outputs in the causal order $p\to o\to a$. Each denoising process can use an independently configured number of steps.

The generated localization $\hat{y}^p$ indicates the relative progress of the robot $\hat{p}_t$ within the current video demonstration window. This relative position is converted into an absolute frame index $\hat{q}_t$ in the full video demonstration, and the next window is centered on this estimate:
{\small
\begin{equation}
\label{eq:window_advance}
  \hat{q}_t=w_t+\lfloor \hat{p}_t L\rfloor,
  \qquad
  w_{t+1}=\hat{q}_t-\lfloor L/2\rfloor .
\end{equation}
}
Here $w_t$ is the start frame index of $\mathcal{W}_t$ in $\tau^d$. The predicted action chunk $\hat{y}^a$ is executed for $H$ steps, after which the loop continues with the updated window $\mathcal{W}_{t+1}$.

\textbf{Persistence across recurring tasks.}
\method acquires and executes each novel task from a single video demonstration $\tau^d$. Each video demonstration is preserved in a demonstration memory $\mathcal{B}$ together with its task instruction and initial scene, enabling a subsequent encounter of the same or a similar task to retrieve it directly without repeated human involvement. The requested task is posed as a query. Its instruction and current scene are compared with the stored entries, and the demonstration with the highest match score above a similarity threshold $\delta$ is retrieved. Otherwise, \method recognizes the task as genuinely novel and requests a new human video.

Each video demonstration $\tau^d_i$ is stored in $\mathcal{B}$ together with its language instruction $\ell_i$, a text embedding $e^{\mathrm{text}}_i \in \mathbb{R}^{d_{\mathrm{emb}}}$ of the instruction, and a scene embedding $e^{\mathrm{img}}_i \in \mathbb{R}^{d_{\mathrm{emb}}}$ of the initial observation. Qwen3-VL-Embedding-8B~\cite{qwen3vl} maps both modalities into a shared $d_{\mathrm{emb}}$-dimensional space with $\ell_2$ normalization:
{\small\begin{equation}\label{eq:memory_embed}
  e^{\mathrm{text}}_i = \frac{g_\psi(\ell_i)}{\|g_\psi(\ell_i)\|_2}, \quad
  e^{\mathrm{img}}_i = \frac{g_\psi(o^d_1)}{\|g_\psi(o^d_1)\|_2},
\end{equation}}
where $g_\psi$ denotes the Qwen3-VL-Embedding-8B encoder and $o^d_1$ is the initial image frame of $\tau^d_i$.

The requested task supplies an instruction $\ell_q$ and the current robot observation $o^r$, encoded into query embeddings $e^{\mathrm{text}}_q$ and $e^{\mathrm{img}}_q$ following Eq.~\ref{eq:memory_embed}. Each stored video demonstration $\tau^d_i$ is then scored by a weighted combination of instruction and scene cosine similarities:
{\small\begin{equation}\label{eq:retrieval_score}
  s_i = \omega \, \cos(e^{\mathrm{text}}_q,\, e^{\mathrm{text}}_i) \;+\; (1 - \omega) \, \cos(e^{\mathrm{img}}_q,\, e^{\mathrm{img}}_i),
\end{equation}}
where $\omega$ balances instruction and scene similarity. The video demonstration with the highest score above $\delta$ is selected and supplied to the inference pipeline as $\tau^d$. Otherwise, \method requests a new human video to execute the task, which is then stored in $\mathcal{B}$ for later reuse.

\bibliographystyle{unsrtnat}
\bibliography{sn-bibliography}

\clearpage
\begin{appendices}

\section*{Supplementary Contents}
\begin{enumerate}
  \item Related work\dotfill\pageref{sec:related_works}
  \begin{enumerate}
    \item[1.1] Learning robot manipulation from demonstrations\dotfill\pageref{sec:related_training}
    \item[1.2] One-shot visual imitation\dotfill\pageref{sec:related_osvi}
    \item[1.3] Temporal alignment and correspondence learning\dotfill\pageref{sec:related_alignment}
  \end{enumerate}
  \item Implementation details\dotfill\pageref{sec:supp_impl}
  \begin{enumerate}
    \item[2.1] Alignment module\dotfill\pageref{sec:supp_align}
    \item[2.2] Autoregressive diffusion model architecture\dotfill\pageref{sec:supp_arch}
    \item[2.3] Training procedure\dotfill\pageref{sec:supp_train}
    \item[2.4] Data processing\dotfill\pageref{sec:supp_data}
    \item[2.5] Inference\dotfill\pageref{sec:supp_infer}
  \end{enumerate}
  \item Notation table\dotfill\pageref{tab:notation}
\end{enumerate}

\clearpage

\section{Related Work}\label{sec:related_works}

\subsection{Learning Robot Manipulation from Demonstrations}\label{sec:related_training}

Learning from demonstrations offers a direct and intuitive way to teach robots new manipulation skills. This section first reviews imitation learning and then discusses two directions that have become prominent in robot learning research, vision--language--action foundation models and world models that incorporate video generation and future prediction.

\subsubsection{Imitation learning from robot and human demonstrations}

Teaching robots to replicate human motor skills is a long-standing objective in robotics~\cite{ArgallSurvey2009, RavichandarSurvey2020}. Imitation learning, or learning from demonstrations (LfD), formulates the problem as recovering a policy from expert state--action trajectories obtained via teleoperation or kinesthetic teaching~\cite{Schaal1999,ZhangLfD2018,Kim2025SRTH}. When the expert and the robot share the same embodiment, behavioral cloning provides an effective framework~\cite{Pomerleau1989}. Recent advances in policy architectures have further improved these approaches. Diffusion Policy~\cite{DiffusionPolicy} models robot action distributions with denoising diffusion processes and achieves strong multimodal behavior coverage. ACT~\cite{ACT} introduces transformer-based action chunking for fine-grained bimanual manipulation on low-cost hardware. These methods have substantially improved the performance attainable through imitation learning when high-quality robot demonstrations are available.

The setting becomes substantially harder when the demonstrator is human, as the embodiment, viewpoint, and dynamics differ from those of the robot. Retargeting methods extract human hand or body poses from video and map them to robot joint configurations through hand-crafted or learned correspondence~\cite{Qin2022DexMV, Wang2024DexCap}, while methods for domain adaptation instead learn embodiment-invariant representations that bridge the visual or state-space gap between human and robot observations~\cite{Sermanet2018TCN, Zakka2022XIRL}. A further line of work leverages human activity directly as a learning signal, learning latent plans or behavioral priors from unstructured human play and in-the-wild video~\cite{PlayLMP,WHIRL,MimicPlay}. Most recently, EgoScale~\cite{EgoScale} establishes a log-linear scaling law between egocentric video data and downstream robot performance, demonstrating that large-scale human video pretraining followed by intermediate training on robot data yields substantial gains with minimal robot supervision~\cite{chen2025graphmimic,chen2026learning,chen2025unifying}.

\subsubsection{Vision--language--action models}

The convergence of large-scale pretraining and robot learning has given rise to vision--language--action (VLA) models that unify perception, language understanding, and robot action generation in a single architecture~\cite{LiIJRR2025}. Early efforts such as Gato~\cite{Gato} demonstrate that a single transformer can be trained across vision, language, and control modalities, albeit with limited manipulation performance. RT-1~\cite{RT1} shows that training a high-capacity transformer on large-scale robot demonstrations yields robust real-world manipulation policies. RT-2~\cite{RT2} extends this approach by jointly fine-tuning a vision--language model on robot action data, which transfers web-scale visual and semantic concepts to robotic control. The Open X-Embodiment initiative~\cite{OpenXEmbodiment} further shows that co-training across diverse embodiments and institutions yields positive transfer, with the resulting RT-X models exhibiting improved performance over single-embodiment baselines.

A broader wave of VLA models has since extended this paradigm across architectures, scales, and embodiments. Octo~\cite{Octo} provides a generalist policy that can be efficiently fine-tuned to new robots and tasks, while OpenVLA~\cite{OpenVLA} offers an open-source VLA trained on the Open X-Embodiment dataset. $\pi_0$~\cite{pi0} introduces flow matching into the VLA framework, achieving state-of-the-art dexterous manipulation and multi-stage task execution, and its successor $\pi_{0.5}$~\cite{pi05} demonstrates open-world generalization to novel environments through large-scale co-training on heterogeneous data, including cross-embodiment robot demonstrations and web data. RDT-1B~\cite{RDT1B} scales diffusion-based policies to one billion parameters for bimanual manipulation, and GR00T N1~\cite{GROOT} provides a foundation model for generalist humanoid robot control through a joint vision--language--action architecture. DexVLA~\cite{DexVLA} introduces a modular diffusion expert plugged into a VLA backbone for dexterous manipulation.

\subsubsection{World models and video generation for robot learning}

World models that predict future states have a long history in robot control. Ha and Schmidhuber~\cite{WorldModels,Ai2025DynamicsModels} demonstrate that recurrent world models learned in latent space can facilitate policy evolution. Dreamer~\cite{Dreamer} learns behaviors by latent imagination, DreamerV2~\cite{DreamerV2} masters Atari with discrete world models, and DreamerV3~\cite{DreamerV3} extends the approach to diverse domains. IRIS~\cite{IRIS} introduces transformer-based world models that achieve strong sample efficiency. In the continuous control setting, PlaNet~\cite{PlaNet} learns latent dynamics for planning from pixels, and TD-MPC2~\cite{TDMPC2} provides scalable and robust latent-space world models.

With the advent of large-scale video generation, pixel-space world models have been explored for manipulation. UniPi~\cite{UniPi} generates future video frames as a planning representation. It converts goals specified in text into visual plans, from which an inverse dynamics model extracts actions. SuSIE~\cite{SuSIE} refines this idea by using a pretrained image-editing diffusion model to generate subgoal images that guide a goal-conditioned policy. Gen2Act~\cite{Gen2Act} generates human demonstration videos in novel scenarios to enable robot generalization, and Dreamitate~\cite{Dreamitate} leverages video generation of demonstrations to learn real-world visuomotor policies. SWIM~\cite{SWIM} pre-trains a world model from human videos and fine-tunes it on robot data, establishing that human video provides useful dynamics priors for manipulation. UniSim~\cite{UniSim} learns an interactive generative simulator from real-world data, enabling policy training within the learned world model.

More recently, world-action models (WAMs) jointly train video prediction and action generation within a shared architecture~\cite{GR1,GR2,LingBotVA,CosmosPolicy,FastWAM,li2026wall}. GR-1~\cite{GR1} pretrains on large-scale video data and fine-tunes for robot action prediction. GR-2~\cite{GR2} extends this to tens of millions of internet video clips, improving generalization. LingBot-VA~\cite{LingBotVA} proposes a Mixture-of-Transformers architecture that jointly predicts future frames and actions in a shared latent space via autoregressive diffusion, using closed-loop rollout for long-horizon dexterous manipulation. Cosmos Policy~\cite{CosmosPolicy} fine-tunes large video generation models as visuomotor robot policies, demonstrating transfer from internet-scale video pretraining. Fast-WAM~\cite{FastWAM} shows that video co-training, rather than explicit future imagination at test time, is the primary source of WAM performance gains. Accordingly, removing test-time video generation yields 4$\times$ faster inference while maintaining competitive performance.

Despite this progress, these methods struggle to generalize beyond the training distribution~\cite{AGNOSTOS}, and learning a new skill typically requires additional task-specific demonstrations and offline training, a cycle that is costly, slow, and risks catastrophic forgetting of previously acquired skills~\cite{CatastrophicForgettingRobot, FLaRe}. \method instead acquires novel manipulation skills at inference time from a single human video demonstration without any parameter update. It achieves 62\% average success on novel tasks and exceeds the strongest zero-shot imitation baseline by 45\%. Compared with supervised fine-tuning on 50 robot demonstrations per task, \method achieves a higher success rate and acquires each skill roughly 507 times faster.

\subsection{One-Shot Visual Imitation}\label{sec:related_osvi}

\subsubsection{One-shot visual imitation methods}

Kuniyoshi et al.~\cite{Kuniyoshi1994} envisioned robots extracting reusable task knowledge from a single observed human performance. This setting, now studied as one-shot visual imitation (OSVI), requires robots to bridge embodiment and viewpoint differences and translate one observed performance into executable robot actions. One line of work constructs modular pipelines that decompose this problem into perception, grounding, and execution stages, estimating the target interaction pose or object configuration from the video demonstration and replaying the resulting trajectory in open loop~\cite{FMimic,chen2024vlmimic,CoarseToFine,OKAMI}. FMimic~\cite{FMimic} leverages vision--language models to extract fine-grained hand--object interaction keypoints from a human video demonstration, then transfers these keypoints to novel scenes via region-to-keypoint mapping and refines execution with pose estimation based on contact. MT3~\cite{MT3} decomposes manipulation trajectories into alignment and interaction phases and retrieves over a thousand tasks from as few as one video demonstration each, assuming the robot grasps each object with the same pose as in the video demonstration. RSRD~\cite{RSRD} recovers articulated 3D part motions from monocular RGB video for bimanual imitation, depending on high-quality part-level 3D reconstruction. Open-loop replay confines these pipelines to reproducing simple demonstrated behaviors and limits the adaptation required for real-world deployment, and independently engineered modules also introduce compounding errors that restrict generalization across diverse manipulation scenarios. 

An alternative line of work seeks to learn a unified model that maps a human video demonstration and current robot observations to robot actions~\cite{FinnOSVI,DAML,ZeroShotVI,TOSVI,MOSAIC,AWDA,DINOBot,Vid2Robot,OSVI-WM,chen2025see}. Early efforts in this direction adopt meta-learning. Finn et al.~\cite{FinnOSVI} propose meta-learning for one-shot visual imitation, and DAML~\cite{DAML} extends this approach to cross-embodiment settings through domain adaptation. Pathak et al.~\cite{ZeroShotVI} demonstrate visual imitation from video without access to action labels. T-OSVI~\cite{TOSVI} uses transformers to capture long-range dependencies in video demonstrations, supporting an embodiment mismatch between the demonstrator and the robot. MOSAIC~\cite{MOSAIC} establishes a more challenging evaluation protocol with completely unseen test tasks, revealing that prior methods struggle to generalize beyond their training distribution. AWDA~\cite{AWDA} improves generalization through attributed waypoints and demonstration augmentation. DINOBot~\cite{DINOBot} leverages vision foundation models for egocentric one-shot imitation via retrieval-based alignment.
More recently, Vid2Robot~\cite{Vid2Robot} encodes the demonstration video via a ViT--Perceiver architecture and conditions a cross-attention transformer policy on the resulting tokens, augmented with temporal cycle-consistency and video--text contrastive losses to align prompt and robot representations. OSVI-WM~\cite{OSVI-WM} introduces an approach guided by a world model, in which a learned causal transformer world model iteratively predicts latent states from concatenated demonstrator and robot representations, decoding the predicted trajectory into physical waypoints. These end-to-end methods eliminate the need for manually designed module interfaces. However, their performance still lags behind that of language-conditioned policies~\cite{pi05} and falls far short of that achieved through task-specific fine-tuning. This gap suggests that learning novel manipulation skills from a single video demonstration remains an open problem.

Existing OSVI methods continue to perform substantially worse than task-specific fine-tuning. This shortfall reflects the structural mismatch between video demonstration and execution. Temporal asynchrony decouples the prediction target from the video demonstration, while differences in embodiment, viewpoint, and appearance make the observed behavior difficult to translate into action. \method resolves this mismatch through \couplemech{} and \stagemech{}. Rather than fixing the prediction target at a temporal offset independent of the video demonstration, \method couples each target to the video demonstration's upcoming content on the \manifold{}, turning the video into the active driver of the robot's predicted actions. Execution is then resolved through \stagemech{}, a causal cascade that first localizes the robot's progress within the video demonstration, then predicts its own future observations, and finally derives actions from these future observations.

\subsection{Temporal Alignment and Correspondence Learning}\label{sec:related_alignment}

\subsubsection{Video temporal alignment methods}

Temporal alignment of videos depicting the same process under different conditions is a fundamental problem in video understanding. The differentiable formulation of Dynamic Time Warping has been a key enabler. Soft-DTW~\cite{SoftDTW} replaces the non-differentiable minimum operator in DTW with a soft-min, producing a differentiable alignment loss that enables gradient-based optimization. D3TW~\cite{D3TW} extends this with a discriminative, margin-based loss for weakly supervised action alignment and segmentation.

A related line of work learns temporal representations in a self-supervised manner, without relying on frame-level correspondence labels. Time-contrastive networks (TCN)~\cite{Sermanet2018TCN} learn viewpoint-invariant representations by enforcing temporal consistency across simultaneously recorded multi-view videos, enabling imitation from observation by matching embeddings across human and robot demonstrations. Temporal cycle-consistency (TCC)~\cite{dwibedi2019tcc} extends this idea to unpaired videos. It learns per-frame embeddings so that a frame mapped to its nearest neighbor in another video and then mapped back returns to its original position. This cycle enables fine-grained alignment without frame-level supervision. Wang~\cite{WangCycleTime} proposes learning spatial correspondence through forward-backward cycle-consistency in time, demonstrating that temporal consistency provides a powerful self-supervisory signal. Hadji et al.~\cite{hadji2021} reformulate DTW as a differentiable probabilistic procedure for path finding. The procedure serves directly as a training objective and extends naturally to a global cycle-consistency loss. It produces monotonic soft matching probabilities for processes that share temporal structure but differ in pace and appearance.

Within robot learning, temporal alignment has been applied to derive reward signals, identify correspondences in task progress, and regularize policy training as an auxiliary loss. XIRL~\cite{Zakka2022XIRL} derives cross-embodiment reward functions from representations learned through temporal alignment, and GraphIRL~\cite{GraphIRL} applies graph abstraction with temporal matching to derive dense reward functions from videos. TCN~\cite{Sermanet2018TCN} uses temporal alignment to identify correspondences in task progress for imitation from observation. Vid2Robot~\cite{Vid2Robot} incorporates temporal cycle-consistency as an auxiliary loss alongside video--text contrastive learning, which aligns prompt and robot representations during policy training.

\method instead uses temporal alignment as a structural mechanism that directly constructs the prediction target, rather than an auxiliary loss or reward signal. A self-supervised alignment module, combining temporal cycle-consistency~\cite{dwibedi2019tcc} with differentiable monotonic path finding~\cite{hadji2021}, recovers frame-level correspondence between the video demonstration and the robot trajectory. Through \couplemech{}, this correspondence redefines the prediction target at each robot timestep. The target is redefined as the segment of the robot's own trajectory that corresponds to the future evolution of the video demonstration. This construction establishes a direct causal dependency between the demonstration content and the training objective, transforming the video from passive context into an active supervisory signal at each step.

\section{Implementation Details}\label{sec:supp_impl}

This section provides the complete implementation details for reproducing \method, covering the alignment module, autoregressive diffusion model architecture, training procedure, data processing, and inference pipeline. Each subsection concludes with a hyperparameter summary table for easy reference.

\subsection{Alignment Module}\label{sec:supp_align}

\textbf{Vision encoder and embedding.}
The alignment module employs Qwen3-VL-Embedding-8B~\cite{qwen3vl} as the shared vision encoder $f_\phi$, loaded in bfloat16 with scaled dot-product attention and gradient checkpointing enabled. The full model is fine-tuned end-to-end during alignment training. Each frame is processed through the complete Qwen3-VL forward pass, and the per-frame representation is extracted from the [EOS] token position in the last hidden layer, yielding a $D{=}1{,}536$-dimensional hidden state. A single learnable linear layer projects the hidden state to a $d{=}128$-dimensional embedding, followed by $\ell_2$ normalization, as formulated in Eq.~\ref{eq:embedding}. The same encoder and projection are applied to both the video demonstration and the robot trajectory.

\textbf{Smooth DTW configuration.}
Smooth Dynamic Time Warping runs forward and backward passes to compute bidirectional alignment. The forward pass accumulates costs along monotonic paths using the smooth minimum operator formulated in Eq.~\ref{eq:sdtw_forward}, with softmin temperature $\gamma{=}1.0$. A column-normalization temperature $\gamma_f{=}0.1$ controls the sharpness of the matching probabilities. When both sequences share the same length, the anti-diagonal parallel implementation reduces the computation from $O(T^2)$ sequential steps to $O(T)$ parallel anti-diagonal sweeps. Boundary conditions initialize the origin cell to its cost value and all border cells to a large constant of $10^9$. The backward table is computed analogously from the terminal cell. The matching probabilities $\beta$ are then obtained by combining the forward and backward tables and applying row-wise softmax normalization, as formulated in Eq.~\ref{eq:sdtw_beta}. The pairwise similarity matrix uses negative squared L2 distance entries scaled by a softmax temperature $\kappa{=}0.1$.

\textbf{Loss configuration.}
The temporal cycle-consistency loss is defined in Eq.~\ref{eq:tcc_loss} and adopts variance-aware mean squared error regression. The predicted cycle-back variance $\nu_i^2$ is clamped with $\epsilon{=}10^{-4}$ for numerical stability under bfloat16, and the variance regularization weight is $\lambda_{\mathrm{var}}{=}0.001$ with label smoothing of $0.1$. The path-cost loss weight is $\lambda_{\mathrm{DTW}}{=}0.3$, normalized by the sum of sequence lengths $T{+}N$, as formulated in Eq.~\ref{eq:d2tw_loss}. Both losses are symmetrized over both alignment directions.

\textbf{Training configuration.}
The module is trained with AdamW, using $\beta_1{=}0.9$, $\beta_2{=}0.999$, $\epsilon{=}10^{-8}$, and weight decay $10^{-5}$. The learning rate is $10^{-5}$ with cosine annealing over 100 warmup steps down to a minimum ratio of $0.3$. Training runs for 10{,}000 steps on 64 GPUs with a per-GPU batch size of 4, yielding an effective batch size of 256. It uses bfloat16 mixed precision, DeepSpeed ZeRO-3, and gradient clipping at $3.0$. Training data comprises same-task trajectory pairs drawn from $\mathcal{M}_{\mathrm{train}}$, spanning both robot--robot and human--robot configurations, with pairs formed by randomly selecting two distinct trajectories of the same task. Data augmentation consists of random horizontal flipping and brightness/contrast jittering.

After training, the alignment module is frozen and used offline to construct the conditioning windows, localization labels, and prediction targets for autoregressive diffusion model training, described in Sec.~\ref{sec:alignment}. At inference time, the alignment module is not invoked. The autoregressive diffusion model autonomously tracks task progress via the learned localization head, described in Sec.~\ref{sec:supp_infer}. Hyperparameters are summarized in Supplementary Table~\ref{tab:hp_align}.

\begin{table}[!htbp]
\centering
\caption{\textbf{Alignment module hyperparameters.}}
\label{tab:hp_align}
\resizebox{\linewidth}{!}{
\begin{tabular}{ll}
\toprule
\textbf{Parameter} & \textbf{Value} \\
\midrule
Vision encoder & Qwen3-VL-Embedding-8B (fine-tuned, bf16) \\
Hidden dimension $D$ / embedding dimension $d_{\mathrm{emb}}$ & 1{,}536 / 128 \\
Embedding projection & Linear + $\ell_2$ normalization \\
SDTW softmin temperature $\gamma$ / column-normalization temperature $\gamma_f$ & 1.0 / 0.1 \\
Softmax temperature $\kappa$ & 0.1 \\
TCC variance weight $\lambda_{\mathrm{var}}$ / label smoothing & 0.001 / 0.1 \\
D2TW loss weight $\lambda_{\mathrm{DTW}}$ & 0.3 \\
Optimizer & AdamW ($\beta_1{=}0.9$, $\beta_2{=}0.999$, $\epsilon{=}10^{-8}$) \\
Learning rate / schedule & $10^{-5}$ / cosine (100 warmup, min ratio 0.3) \\
Weight decay / gradient clipping & $10^{-5}$ / 3.0 \\
Training steps / GPUs / batch per GPU & 10{,}000 / 64 / 4 \\
Precision / distributed strategy & bfloat16 / DeepSpeed ZeRO-3 \\
\bottomrule
\end{tabular}
}
\end{table}

\subsection{Autoregressive Diffusion Model Architecture}\label{sec:supp_arch}

\textbf{VAE.}
Robot visual observations and video demonstration frames are encoded by WanVideoVAE~\cite{wan2025} with 48 latent channels and $8{\times}$ spatial downsampling. Multi-view images from all three cameras are vertically concatenated in pixel space prior to VAE encoding, producing a combined frame of height $224{\times}3{=}672$ pixels that is mapped to an $84{\times}28$ latent grid per timestep. The VAE is frozen throughout training.

\textbf{\Videoexpert.}
The \videoexpert{} is a 30-layer Diffusion Transformer~\cite{DiT} initialized from Wan2.2-TI2V-5B~\cite{wan2025}, with hidden dimension 3{,}072, feed-forward dimension 14{,}336, 24 attention heads at 128 dimensions per head, and patch size $[1, 2, 2]$. It processes the video demonstration tokens together with the current robot observation and future observation tokens, so its predicted future observations remain grounded in the current robot state. A causal attention mask on the first frame lets the first frame, the current robot observation, attend only to its own tokens. All subsequent frames instead attend bidirectionally to all frames, including the first. This design prevents the deterministic current observation from being contaminated by noisy future observation tokens, while still letting the future observation tokens condition on the current observation.

Each transformer block employs Adaptive Layer Normalization (AdaLN) with six learned modulation components per layer. These components comprise shift, scale, and gate for both self-attention and feed-forward sub-layers. The diffusion timestep is embedded via sinusoidal encoding, followed by a two-layer MLP that maps freq\_dim$\,{=}\,$256 to hidden\_dim. It is then projected to the six modulation parameters through a SiLU-activated linear layer. Separated timestep embedding is used. The first frame always receives a clean timestep of zero, while subsequent frames receive the actual diffusion timestep. This reflects the asymmetry between the deterministic input observation and the noisy generation targets.

Three-dimensional rotary position embedding (3D RoPE) encodes spatial and temporal structure. Separate frequency components for the temporal, height, and width axes are precomputed and applied to query and key vectors via complex rotation. Video demonstration frames receive temporal positions $[0, 1, \ldots, f_d{-}1]$, robot observation frames continue from $[f_d, \ldots, f_d{+}f_r{-}1]$, and the localization token receives a dedicated temporal index one past all frames with zero spatial coordinates.

\textbf{Video demonstration conditioning.}
Video demonstration tokens $z^d$ are placed at the beginning of the self-attention sequence, as formulated in Eq.~\ref{eq:sequence}. They participate in the shared self-attention computation alongside robot observation and action tokens. The attention mask enforces unidirectional conditioning. Video demonstration tokens attend only to other video demonstration tokens, through full self-attention within the video demonstration window, while robot observation and action tokens can attend to the video demonstration tokens. This design allows the model to extract information from the video demonstration without it being influenced by the current robot state. When the video demonstration is dropped during training, with probability $0.5$, the corresponding tokens are replaced with zero padding and masked out of all attention computations. The text description then serves as the sole task conditioning signal.

Headwise gating modulates the influence of video demonstration information at each attention head. Gate parameters are initialized with zero weights and a bias of $5.0$, starting the model with approximately neutral, open gating and allowing task-specific attention patterns to emerge during training.

\textbf{\Actionexpert.}
The \actionexpert{} is a 30-layer DiT with a hidden dimension of 1{,}024, a feed-forward dimension of 4{,}096, and the same attention-head configuration as the \videoexpert{}. Both use 24 heads with 128 dimensions per head. This shared configuration permits shared self-attention in the MoT framework. Its query, key, and value projections map the 1{,}024-dimensional hidden state to this shared 3{,}072-dimensional attention space. The attention output is then projected back to 1{,}024 dimensions before the feed-forward sub-layer. Each action step is independently encoded by a linear projection from the action dimension, mapping $\mathbb{R}^{20}$ to $\mathbb{R}^{1024}$. This yields one token per timestep, for a total of $H{=}32$ action tokens. The \actionexpert{} applies one-dimensional RoPE over the action sequence.

The backbone parameters of the \actionexpert{} are initialized from the \videoexpert{} via dimension-wise linear interpolation. For each weight tensor whose last dimension differs between the two experts, the pretrained weights are linearly interpolated from $d_v{=}3{,}072$ to $d_a{=}1{,}024$ along that dimension. They are then scaled by $\alpha{=}\sqrt{d_v/d_a}$ to preserve the variance of activations despite the dimensional reduction. Action-specific parameters that have no counterpart in the \videoexpert{}, such as the input projection and output head, are initialized with standard Kaiming uniform initialization. Headwise gating is applied to the self-attention in the \actionexpert{}, with the same zero-weight, bias-$5.0$ initialization.

\textbf{Mixture-of-Transformers.}
In the MoT architecture, formulated in Eq.~\ref{eq:mot_attn}, the two experts share self-attention by concatenating their query, key, and value projections at each layer. The joint attention output is then split by expert and processed by branch-specific feed-forward networks. Gradient checkpointing is applied to the mixed attention computation during training to reduce memory consumption.

\textbf{Text and proprioceptive conditioning.}
The cross-attention conditioning sequence $c$ is formulated in Eq.~\ref{eq:context}. It comprises text embeddings encoded by UMT5-XXL~\cite{umt5} at 4{,}096 dimensions, with a context length of 512 and a maximum tokenizer length of 512. It also comprises the proprioceptive state, projected via a learnable linear layer from $\mathbb{R}^{20}$ to $\mathbb{R}^{4096}$. Text embeddings are computed offline using the frozen UMT5-XXL encoder and loaded during training. Both expert branches receive $c$ via cross-attention at every transformer layer. The framework accepts either a language instruction, a video demonstration, or both as task specification at inference time.

\textbf{Localization token.}
The scalar localization value $p_t$ is projected to the hidden dimension of the \videoexpert{} by a learnable encoder $E_p$, a two-layer MLP with GELU activation. It maps $\mathbb{R}^{1}$ to $\mathbb{R}^{3072}$ and then to $\mathbb{R}^{3072}$, producing a single localization token. The decoder $D_p$ maps the generated representation back to a scalar prediction through LayerNorm followed by a two-layer MLP with GELU activation. This MLP maps $\mathbb{R}^{3072}$ to $\mathbb{R}^{3072}$ and then to $\mathbb{R}^{1}$.

To bridge the gap between ground-truth conditioning at training time and model-predicted conditioning at inference, the clean progress token $y^p$ is augmented with bounded Gaussian noise during training. This token conditions subsequent generation targets, and the noise has standard deviation $0.5$, applied with probability $0.5$. The noised value is then clamped to $[-1, 1]$. This noisy clean progress (NCP) scheme improves the robustness of autoregressive conditioning. Progress denoising uses a dedicated flow matching scheduler with shift$\,{=}\,3.0$.

\textbf{Autoregressive attention mask.}
The causal ordering among the three generation targets, described in Sec.~\ref{sec:wam}, is enforced through the self-attention mask. The noisy localization token can attend to $z^d$ and $z^o$ but not to subsequent targets. The noisy observation tokens can attend to $z^d$, $z^o$, and the clean localization token, but not to action tokens. The noisy action tokens can attend to $z^d$, $z^o$, the clean localization token, and the clean observation tokens. During training, clean tokens are populated with ground-truth values and their gradients are stopped so that each target's loss does not back-propagate into preceding targets. Hyperparameters are summarized in Supplementary Table~\ref{tab:hp_arch}.

\begin{table}[!htbp]
\centering
\caption{\textbf{Autoregressive diffusion model architecture hyperparameters.}}
\label{tab:hp_arch}
\resizebox{\linewidth}{!}{
\begin{tabular}{lll}
\toprule
\textbf{Module} & \textbf{Parameter} & \textbf{Value} \\
\midrule
\multicolumn{3}{l}{\textit{\Videoexpert}} \\
& Backbone & Wan2.2-TI2V-5B (5B parameters) \\
& Layers / hidden dim / FFN dim & 30 / 3{,}072 / 14{,}336 \\
& Attention heads / head dim & 24 / 128 \\
& Patch size & $[1, 2, 2]$ \\
& VAE latent channels / spatial downsample & 48 / $8{\times}$ \\
& Attention mask & Causal on first frame \\
& Positional encoding & 3D RoPE (temporal, height, width) \\
& Modulation & AdaLN (6 components per layer) \\
& Timestep embedding & Separated (first frame $\sigma{=}0$) \\
\midrule
\multicolumn{3}{l}{\textit{\Actionexpert}} \\
& Layers / hidden dim / FFN dim & 30 / 1{,}024 / 4{,}096 \\
& Attention heads / head dim & 24 / 128 \\
& Action tokenization & Linear($\mathbb{R}^{20} \to \mathbb{R}^{1024}$), one token per step \\
& Initialization & Linear interpolation from \videoexpert{} ($\alpha{=}\sqrt{d_v/d_a}$) \\
& Positional encoding & 1D RoPE \\
\midrule
\multicolumn{3}{l}{\textit{Conditioning and gating}} \\
& Text encoder & UMT5-XXL (offline, 4{,}096-dim, context length 512) \\
& Proprioceptive projection & Linear($\mathbb{R}^{20} \to \mathbb{R}^{4096}$) \\
& Video demonstration conditioning & Self-attention \\
& Headwise gate initialization & Weights$\,{=}\,0$, bias$\,{=}\,5.0$ \\
& Video demonstration drop probability & 0.5 \\
& Progress encoder $E_p$ & MLP($1 \to 3072 \to 3072$, GELU) \\
& Progress decoder $D_p$ & LayerNorm $\to$ MLP($3072 \to 3072 \to 1$, GELU) \\
& NCP noise scale / probability & 0.5 / 0.5 \\
\bottomrule
\end{tabular}
}
\end{table}

\subsection{Training Procedure}\label{sec:supp_train}

Stage 1 performs same-embodiment pretraining, and Stage 2 performs human--robot adaptation. Both stages share an identical training configuration with full-parameter updates. The optimizer is AdamW with $\beta_1{=}0.9$, $\beta_2{=}0.95$, and weight decay $10^{-2}$. Training uses a constant learning rate of $10^{-5}$. Training is distributed across 64 GPUs with a per-GPU batch size of 2, yielding an effective batch size of 128. It uses bfloat16 mixed precision and DeepSpeed ZeRO Stage~1. Gradients are clipped to a maximum norm of $1.0$. No exponential moving average (EMA) of model weights is used.

The loss weights are $\lambda_o{=}1.0$, $\lambda_a{=}10.0$, and $\lambda_p{=}1.0$, as formulated in Eq.~\ref{eq:total_loss}. The elevated action loss weight compensates for the lower token count of the action sequence relative to the observation token grid and ensures balanced gradient magnitudes across the two expert branches.

\textbf{Flow matching configuration.}
Flow matching uses 1{,}000 training timesteps with a shift of $5.0$ for both observation and action schedulers, and a shift of $3.0$ for the progress scheduler. All three targets follow the same velocity-prediction parameterization, as formulated in Eq.~\ref{eq:flow_matching}. Timesteps $\sigma$ are sampled by drawing $u \sim \mathrm{Uniform}[0,1]$ and applying the shifted schedule $\sigma = \phi(u, s) = s \cdot u \,/\, (1 + (s{-}1) \cdot u)$, where $s$ is the shift parameter. A Gaussian-weighted importance function $w(\sigma) = \exp\!\big({-}2\big((\sigma - \sigma_{\max}/2)\,/\,\sigma_{\max}\big)^2\big)$, centered at the midpoint of the schedule, concentrates training signal on intermediate noise levels where the denoising task is most informative.

\textbf{Training scale.}
Stage~1 trains for 500{,}000 steps on same-embodiment robot--robot pairs formed from 193{,}462 robot trajectories spanning 229 tasks. Stage~2 trains for 100{,}000 steps on human--robot pairs formed from an additional 5{,}847 self-collected human video demonstrations, each paired with robot trajectories of the same task drawn from that corpus. Hyperparameters are summarized in Supplementary Table~\ref{tab:hp_train}.

\begin{table}[!htbp]
\centering
\caption{\textbf{Training hyperparameters (shared by Stage~1 and Stage~2).}}
\label{tab:hp_train}
\resizebox{\linewidth}{!}{
\begin{tabular}{ll}
\toprule
\textbf{Parameter} & \textbf{Value} \\
\midrule
Optimizer & AdamW ($\beta_1{=}0.9$, $\beta_2{=}0.95$) \\
Learning rate / schedule & $10^{-5}$ / constant \\
Weight decay / max gradient norm & $10^{-2}$ / 1.0 \\
GPUs / batch per GPU / effective batch & 64 / 2 / 128 \\
Precision / distributed strategy & bfloat16 / DeepSpeed ZeRO-1 \\
EMA & None \\
Loss weights $\lambda_o / \lambda_a / \lambda_p$ & 1.0 / 10.0 / 1.0 \\
Flow matching timesteps & 1{,}000 \\
Shift (observation, action / progress) & 5.0 / 3.0 \\
Prediction type (all targets) & Velocity (Eq.~\ref{eq:flow_matching}) \\
Timestep sampling & Shifted schedule + Gaussian weighting \\
Stage 1 steps / data & 500K / 193{,}462 robot trajectories (229 tasks) \\
Stage 2 steps / data & 100K / 5{,}847 human--robot pairs \\
\bottomrule
\end{tabular}
}
\end{table}

\subsection{Data Processing}\label{sec:supp_data}

\textbf{Observations.}
Three RGB cameras observe the workspace, including one wrist-mounted camera on each arm and one camera providing a static third-person view. Each camera captures $224{\times}224$ images, resized by aligning dimensions to a multiple of 8 while preserving aspect ratio. Multi-view observations are vertically concatenated in pixel space before VAE encoding, producing a combined frame of height $224 \times 3 = 672$ pixels. The robot history length is $K{=}0$, meaning only the current frame is used and no historical frames are included.

\textbf{Actions and proprioception.}
Each arm operates in a 10-dimensional action space, comprising a 3-dimensional Cartesian position, a 6-dimensional rotation representation~\cite{zhou2019continuity}, and a 1-dimensional gripper aperture. Together, the two arms form a 20-dimensional bimanual action space. The proprioceptive state is 20-dimensional, encoding the end-effector pose of both arms. Actions are normalized per-dimension to $[-1, 1]$ via linear min-max scaling computed from training statistics. Six-dimensional rotation representations are used throughout training and converted to Euler angles through the rotation matrix at inference time.

\textbf{Temporal structure.}
Each video demonstration window $\mathcal{W}$ contains $L{=}6{\times}32{=}192$ raw video demonstration frames covering the coupled segment and its surrounding context. Temporal downsampling at an action-to-video ratio of $8{:}1$ produces the $192/8{=}24$ video frames provided to the model. The mapping derived from alignment and formulated in Eq.~\ref{eq:pred_target} establishes a monotonic correspondence at the frame level between video demonstration frames and robot frames. The resulting target is defined at action resolution and contains $H{=}32$ aligned robot timesteps per cycle. All $H$ actions are retained for action prediction, while the corresponding observations are uniformly downsampled at an $8{:}1$ ratio before video encoding, yielding $H/8{=}4$ target frames for video prediction. During training, the video demonstration window position is randomly offset to vary the surrounding context. The base action frame interval is 15 frames at 32\,Hz, and it is randomly scaled by a factor drawn uniformly from $[0.5, 2.0]$ to improve robustness to execution speed variation. Static frames, identified by rotation, translation, and gripper displacement thresholds of $0.001$, are removed from trajectories before sampling.

\textbf{Data augmentation.}
Augmentation is applied per-sample with temporally consistent parameters across all frames. Brightness, contrast, and saturation are each scaled by a factor drawn uniformly from $[0.7, 1.3]$, and hue is shifted uniformly in $[-0.05, 0.05]$. Gaussian blur, with $\sigma$ drawn from $[0.1, 2.0]$, is applied with probability $0.5$. Random cropping with scale in $[0.8, 1.0]$ is also applied. Hyperparameters are summarized in Supplementary Table~\ref{tab:hp_data}.

\begin{table}[!htbp]
\centering
\caption{\textbf{Data processing hyperparameters.}}
\label{tab:hp_data}
\resizebox{\linewidth}{!}{
\begin{tabular}{ll}
\toprule
\textbf{Parameter} & \textbf{Value} \\
\midrule
Image resolution / resize & $224 \times 224$ / resize aligned to multiples of 8 \\
Number of cameras & 3 (2 wrist + 1 third-person) \\
Multi-view encoding & Vertical concatenation in pixel space before VAE \\
Robot history $K$ & 0 \\
Action dimension / proprioceptive dimension & 20 / 20 (EEF) \\
Action normalization & Linear min-max to $[-1,1]$ \\
Video demonstration window length $L$ / action chunk $H$ & 192 frames ($6{\times}32$) / 32 steps \\
Action-to-video downsample ratio & $8{:}1$ \\
Action frame interval / random scale & 15 / $[0.5, 2.0]$ \\
Static frame thresholds (rotation/translation/gripper) & 0.001 / 0.001 / 0.001 \\
\bottomrule
\end{tabular}
}
\end{table}

\subsection{Inference}\label{sec:supp_infer}

At each step, the three targets are generated through sequential autoregressive denoising (Eq.~\ref{eq:ar_inference}), using a first-order Euler ODE solver. Localization is denoised in 10 steps from pure Gaussian noise, producing the clean progress token $\hat{y}^p$ that conditions subsequent stages. Visual prediction is then denoised in 20 steps, generating the predicted future observations $\hat{y}^o$ conditioned on $\hat{y}^p$. Finally, action is denoised in 20 steps, producing the action chunk $\hat{y}^a$ conditioned on both $\hat{y}^p$ and $\hat{y}^o$. Each stage uses its respective scheduler configuration, with shift$\,{=}\,3.0$ for localization and shift$\,{=}\,5.0$ for visual prediction and action. The inference schedule samples uniform steps from $\sigma{=}1$ to $\sigma{=}0$ and transforms them through the same shifted schedule $\phi$ used during training.

The predicted action chunk of $H{=}32$ steps is fully executed before the next step. The video demonstration window $\mathcal{W}$ is advanced via Eq.~\ref{eq:window_advance} with the predicted localization scalar, and the window start is monotonically clipped to $[w_t,\,|\tau^d|{-}L]$ to prevent backward drift. Hyperparameters are summarized in Supplementary Table~\ref{tab:hp_infer}.

\begin{table}[!htbp]
\centering
\caption{\textbf{Inference hyperparameters.}}
\label{tab:hp_infer}
\begin{tabular}{ll}
\toprule
\textbf{Parameter} & \textbf{Value} \\
\midrule
ODE solver & First-order Euler \\
Localization denoising steps & 10 \\
Observation denoising steps & 20 \\
Action denoising steps & 20 \\
Action execution & Full chunk (32 steps) \\
\bottomrule
\end{tabular}
\end{table}

\begin{table}[!htbp]
\caption{Summary of notation used in the main text.}
\label{tab:notation}
\centering
\footnotesize
\begin{tabular}{@{}ll@{}}
\toprule
\textbf{Symbol} & \textbf{Description} \\
\midrule
\multicolumn{2}{@{}l}{\textit{Tasks and trajectories}} \\
$\mathcal{M}$, $\mathcal{M}_{\mathrm{train}}$, $\mathcal{M}_{\mathrm{test}}$ & Task set, training subset, testing subset \\
$m$ & Individual manipulation task \\
$\tau^d$ & Video demonstration \\
$\tau^r$ & Robot trajectory \\
$o^d_t \in \mathcal{O}_d$, $o^r_t \in \mathcal{O}_r$ & video demonstration / robot observation at time $t$ \\
$s^r_t \in \mathcal{S}$ & Robot proprioceptive state at time $t$ \\
$a_t \in \mathcal{A}$ & Robot action at time $t$ \\
$T_i$, $N_j$ & Length of video demonstration $i$ / robot trajectory $j$ \\
\midrule
\multicolumn{2}{@{}l}{\textit{Embeddings and representations}} \\
$f_\phi$ & Shared vision encoder \\
$h_t$ & Per-frame representation ($D$-dimensional) \\
$P$ & Learnable projection matrix \\
$\mathbf{e}_t$ & Frame embedding ($d_{\mathrm{emb}}$-dimensional) \\
$\mathbf{d}_t$, $\mathbf{r}_t$ & Video demonstration / robot frame embeddings \\
$\mathbf{d}$, $\mathbf{r}$ & Video demonstration / robot embedding sequences \\
\midrule
\multicolumn{2}{@{}l}{\textit{Temporal alignment}} \\
$S$ & Pairwise distance-based similarity matrix \\
$\kappa$ & Temperature parameter \\
$\beta^{d \to r}$, $\beta^{r \to d}$ & Soft matching probabilities \\
$R(i,j)$, $E(i,j)$ & Forward / backward DTW tables \\
$\gamma$ & Smoothness parameter for smooth minimum \\
$\gamma_f$ & Column-normalization temperature \\
$\tilde{\mathbf{r}}_i$ & Soft nearest neighbor (robot) \\
$\hat{\beta}_{ik}$ & Cycle-back distribution \\
$\mu_i$, $\nu_i^2$ & Cycle-back mean / variance \\
$\pi^{r \to d}$, $\pi^{d \to r}$ & Monotonic frame-level mappings \\
\midrule
\multicolumn{2}{@{}l}{\textit{Prediction target and conditioning}} \\
$H$ & Number of action steps in the prediction target \\
$t_i$ & Robot timestep in the prediction target \\
$\mathcal{T}_t$ & Prediction target at robot timestep $t$ \\
$\mathcal{T}_t^o$, $\mathcal{T}_t^a$ & Observation / action components of $\mathcal{T}_t$ \\
$\mathcal{W}$ & Video demonstration sliding window \\
$L$ & Window length \\
$p_t$ & Localization value (relative position within $\mathcal{W}$) \\
$w_t$ & Start frame index of $\mathcal{W}$ in $\tau^d$ \\
\midrule
\multicolumn{2}{@{}l}{\textit{Model and diffusion}} \\
$\pi_\theta$ & Policy with parameters $\theta$ \\
$\hat{v}_p$, $\hat{v}_o$, $\hat{v}_a$ & Predicted localization / observation / action velocity \\
$z^d$, $z^o$ & Video demonstration / robot observation tokens \\
$y^p$ & Localization token \\
$y^o$, $y^a$ & Target observation / action targets \\
$\epsilon$ & Gaussian noise vector \\
$\sigma$ & Diffusion timestep \\
$y^p_{\sigma}$, $y^o_{\sigma}$, $y^a_{\sigma}$ & Noisy localization / observation / action tokens \\
$E_p$ & Localization encoder \\
$M_s$ & Proprioceptive projection matrix \\
$\ell$ & Language instruction \\
$c$ & Cross-attention conditioning sequence \\
\midrule
\multicolumn{2}{@{}l}{\textit{Loss functions}} \\
$\mathcal{L}_{\mathrm{TCC}}$ & Temporal cycle-consistency loss \\
$\mathcal{L}_{\mathrm{DTW}}$ & Path-cost loss \\
$\mathcal{L}_{\mathrm{align}}$ & Total alignment loss \\
$\mathcal{L}_{\mathrm{obs}}$, $\mathcal{L}_{\mathrm{act}}$, $\mathcal{L}_{\mathrm{loc}}$ & Observation / action / localization loss \\
$\mathcal{L}$ & Total training loss \\
$\lambda$, $\lambda_{\mathrm{DTW}}$, $\lambda_o$, $\lambda_a$, $\lambda_p$ & Loss weights \\
\midrule
\multicolumn{2}{@{}l}{\textit{Inference}} \\
$\hat{y}^p$, $\hat{y}^o$, $\hat{y}^a$ & Predicted localization / observation / action \\
$\hat{p}_t$ & Decoded localization scalar \\
$\hat{q}_t$ & Absolute frame index in the video demonstration \\
\bottomrule
\end{tabular}
\end{table}

\end{appendices}

\end{document}